\documentclass[lettersize,journal]{IEEEtran}
\pdfoutput=1
\usepackage{tikz}
\usetikzlibrary{positioning, arrows}
\usetikzlibrary{matrix}
\usepackage{amsmath,amsfonts}
\usepackage{algorithmic}
\usepackage{algorithm}
\usepackage{array}
\usepackage[caption=false,font=normalsize,labelfont=sf,textfont=sf]{subfig}
\usepackage{textcomp}
\usepackage{url}
\usepackage{verbatim}
\usepackage{dblfloatfix}
\usepackage{graphicx}
\usepackage{subfig}
\usepackage{cite}
\usepackage{amssymb}
\usepackage{multirow}
\usepackage{tabularx}
\usepackage{xcolor}
    \newcolumntype{L}{>{\raggedright\arraybackslash}X}
\newcommand\blfootnote[1]{%
  \begingroup
  \renewcommand\thefootnote{}\footnote{#1}%
  \addtocounter{footnote}{-1}%
  \endgroup
}
\begin{document}

\title{Bandwidth Reservation for Time-Critical Vehicular Applications: A Multi-Operator Environment

}

\author{Abdullah Al-Khatib, Abdullah Ahmed, Klaus Moessner, Holger Timinger}

\maketitle

\begin{abstract}
Onsite bandwidth reservation requests often face challenges such as price fluctuations and fairness issues due to unpredictable bandwidth availability and stringent latency requirements. Requesting bandwidth in advance can mitigate the impact of these fluctuations and ensure timely access to critical resources. In a multi-Mobile Network Operator (MNO) environment, vehicles need to select cost-effective and reliable resources for their safety-critical applications. This research aims to minimize resource costs by finding the best price among multiple MNOs. It formulates multi-operator scenarios as a Markov Decision Process (MDP), utilizing a Deep Reinforcement Learning (DRL) algorithm, specifically Dueling Deep Q-Learning. For efficient and stable learning, we propose a novel area-wise approach and an adaptive MDP synthetic close to the real environment. The Temporal Fusion Transformer (TFT) is used to handle time-dependent data and model training. Furthermore, the research leverages Amazon spot price data and adopts a multi-phase training approach, involving initial training on synthetic data, followed by real-world data. These phases enable the DRL agent to make informed decisions using insights from historical data and real-time observations. The results show that our model leads to significant cost reductions, up to 40\%, compared to scenarios without a policy model in such a complex environment.

\end{abstract}


\section{Introduction}
\IEEEPARstart{W}{\lowercase{ith}} the advent of the Internet of Things (IoT) era, the number of connected vehicles has significantly increased, inspiring a wide range of innovative applications such as autonomous driving \cite{liu2018distributed, du2017computation, yuan2018toward}.  These applications are safety-critical and time-sensitive, typically driven by complex real-time computations, posing significant challenges for resource- and capability-constrained environments \cite{wang2019cooperative}. In addition, limited resources can further hinder real-time processing on board vehicles \cite{wang2020architectural}. \blfootnote {
Abdullah Al-Khatib, Abdullah Ahmed and Holger Timinger are with the Institute for Data and Process Science, Landshut University of Applied Sciences, Germany (e-mail: {Abdullah.Al-Khatib, s-aahmed, Holger.Timinger}@haw-landshut.de).

Klaus Moessner is with the Professorship for Communications Engineering, Technical University Chemnitz, Germany (e-mail: Klaus.Moessner@etit.tu-chemnitz.de).}To address this, application data can be offloaded to centralized cloud servers or edge cloud servers via 5G Vehicle-to-Infrastructure (V2I) connections \cite{jiang2019toward}, \cite{chiang2016fog}. The primary challenge lies in meeting stringent requirements, particularly in communication/network resources (bandwidth), which require ultra-low-latency and ultra-reliable network connectivity with deterministic, guaranteed access to computing resources at the nearby edge (via access points such as roadside units and base stations) \cite{palattella2019fog}. This aspect is often overlooked in most of the literature when discussing the context of edge resources \cite{liwang2021resource, liwang2022overbooking, qi2023matching}. As bandwidth is a scarce resource, it is often prioritized as the first resource to be utilized when the offloading process begins, and it has complex requirements, as mentioned earlier \cite{al2022optimal, al2022heuristic}. Therefore, the development of cost-effective and responsive resource provisioning techniques becomes critical to ensure the necessary bandwidth resources for the aforementioned computation-intensive vehicular applications.

Reservation approaches, which provide guaranteed access to scarce resources, have received considerable attention in research \cite{sciancalepore2017mobile, al2020priority, al2021bandwidth, al2022optimal, al2022heuristic}. However, the majority of existing studies on reservation mechanisms focus on the protocol perspective of network-side reservations \cite{liwang2022overbooking, sciancalepore2017mobile, al2020priority, al2021bandwidth}. In contrast, individual reservation schemes \cite{niyato2008competitive, zang2019filling, chen2020edge, zang2021soar, al2022optimal, zhang2023dynamic, al2024optimizing, al2024blockchain, al2024resources} take the side of the resource consumer and focus on the economic aspect of minimizing user expenditure. These schemes allow resource consumers (i.e., vehicles) to place individual reservation requests, after which an MNO allocates the resources based on these requests. A vehicle might reserve specific bandwidth and computational capabilities to meet its real-time processing needs for a time-sensitive vehicular application. However, there are very few individual reservation schemes available that provide the efficient and cost-effective reservations required for these applications.

From a commercial perspective, MNOs charge for resource allocation using several traditional pricing models. These models include Pay-As-You-Go (PAYG) \cite{AmazonW} or long-term upfront subscription fees for reserving future resources \cite{Atandt}. This discussion includes various pricing schemes employed by different MNOs, such as AT\&T, Verizon, MTN, Ericsson, Vodafone, and cloud/edge computing providers such as Amazon Web Service (AWS), Microsoft Azure, and AppNexus. Recently, dynamic pricing has emerged as a promising solution that is receiving attention from both academia and industry in the context of resource management in edge computing. It effectively manages peak-time periods, aiming to ensure sufficient Quality of Service (QoS) for time-critical applications \cite{tasiopoulos2019fogspot, liao2021intelligent, luong2016data}. However, existing solutions for individual reservation schemes typically evaluate their options only with PAYG and subscription or reserve models, while often neglecting the dynamic pricing set by MNOs \cite{tasiopoulos2019fogspot, al2022optimal}. As a result, vehicles face various challenges, such as the timing of place reservation requests, leading to higher costs or missed opportunities for cost savings in this dynamic environment. The adoption of this pricing approach has prompted MNOs to reconsider their purchasing programs to yield more revenue \cite{xu2013dynamic}. Take some real-world companies, such as AWS, which offers spot pricing for virtual instances, and MTN, China Telecom, and Uninor, which have offered time-dependent pricing for bandwidth resources. Where the price dynamically changes after one hour or even one minute to achieve a balance between supply and demand \cite{sen2013survey}. Dynamic pricing is a growing field in revenue management, with applications in industries like smart grid \cite{ferdous2017optimal} and spectrum trading \cite{yang2012pricing}. Recent studies \cite{al2022optimal, al2022heuristic} show how vehicles can minimize bandwidth costs and risks associated with dynamic pricing and resource unavailability from single MNOs by carefully timing their requests. 


In a multi-operator environment, when a single MNO is unable to cover one road segment, another MNO may provide coverage for the remaining part of the road segment to ensure a connection for the vehicle. For example, a vehicle needs to maintain connectivity with MNOs along the entire road segment from the time of entering the segment, denoted as $\tau_{\text{start}}$, to the time of exiting it, denoted as $\tau_{\text{end}}$, as shown in Fig. \ref{fig:system_model}. Notably, in the sub-regions between $(\tau_{k(1)}, \tau_{w(1)})$ and $(\tau_{k(5)}, \tau_{w(5)})$, where each $\tau_{k(i\in I)}$ is the time of entering the sub-segment and $\tau_{w(i\in I)}$ is the time of exiting the sub-segment, the vehicle lacks the option of connecting to a single MNO. Similarly, in the regions between $(\tau_{k(2)}, \tau_{w(2)})$, $(\tau_{k(3)}, \tau_{w(3)})$, and $(\tau_{k(4)}, \tau_{w(4)})$, where overlapping coverages occur, the vehicle is under the coverage of multiple MNOs. The vehicle faces the task of selecting the most cost-effective option, which is the focus of this paper. Furthermore, the objective is not only to select the MNO with the lowest cost but also to optimize the time over all road segments.

In the literature, numerous existing works have been dedicated to studying onsite reservation request problems in a single MNO scenario \cite{niyato2008competitive, chen2020edge, zang2019filling,al2022optimal, zang2021soar, al2024optimizing, al2024blockchain, al2024resources, gu2016efficient, cao2016share, chen2020stackelberg}. However, few of these studies have considered economic issues with different pricing strategies. In complex multi-operator scenarios, the optimal price often depends on multiple factors and interactions. These include dynamic and varied pricing strategies by MNOs, as well as the number of available MNOs and the overlapping areas between the MNOs in a certain road segment. Therefore, advanced machine learning algorithms, especially DRL, have attracted the attention of academia and industry, gaining significant popularity for navigating complexity and making informed decisions by considering various parameters.
Given that the prices offered by a MNO are not necessarily derived from a stationary distribution, it is posited that deep learning models consistently exhibit superior performance compared to traditional models, such as ARIMA, SARIMA and Exponential Smoothing (ETS) \cite{lara2021experimental}. The inherent non-stationarity of price dynamics justifies the adaptability and flexibility offered by deep learning methods.
Moreover, in the context of the reservation problem amid multiple MNOs, the scenario can be aptly modeled as an instantiating of the well-known secretary problem, as it can be formulated in terms of the Bellman equation. In this regard, the employment of a Q-learning model is deemed particularly fitting \cite{sutton2018reinforcement}.

Despite the benefits of implementing DRL as a control method for reservations, there is one major drawback to training the DRL agent. Specifically, when a DRL agent is trained in a real-world environment, huge costs will arise. Therefore, creating an environment that simulates the MNOs reservation scenario for training the DRL becomes necessary.
Since the DRL model trained for one environment can be employed in another without guarantee of delivering the same performance, fine-tuning the model for the new environment will be crucial. This fine-tuning serves as a means of employing transfer learning to enhance its performance.
The contributions of this paper can be summarized as follows:

\begin{itemize} \item The paper introduces a well-defined problem formulation addressing bandwidth reservation challenges in a multi-operator setting with a focus on cost minimization. The problem is structured as a Markov Decision Process (MDP), considering the time variability of MNO prices. To handle the complexities of a multi-operator environment, a structured framework is proposed to capture prices over different time-defined cost coverage areas.

\item A novel area-wise approach is explored, which provides a robust approximation for pricing in a multi-MNO environment. The strategy optimizes the Q-learning process by presenting individual areas to the agent during training episodes, resulting in improved computational efficiency and adaptability.

\item The integration of Dueling DQN with TFT is proposed to enhance decision-making in a time-series context. Dueling DQN efficiently manages discrete action spaces relevant to resource allocation, while TFT enables effective feature extraction from complex, multi-dimensional time-series data, addressing challenges such as temporal trends, seasonality, and data sparsity.

\item A scalable and practical multi-MNO framework is developed, explicitly validated across multiple configurations, datasets, and regions. This ensures adaptability and generalization to real-world vehicular network setups while maintaining computational efficiency.

\item A multi-phase training algorithm is introduced to improve the performance of Dueling DQN for bandwidth reservation. This algorithm leverages synthetic, fine-tuned, and real-world data during training to optimize decision-making under dynamic conditions.

\item The proposed system prioritizes interpretability and practicality by optimizing discrete action spaces, simplifying model complexity, and ensuring transparency in decision-making through Q-value outputs. These features address real-world concerns in multi-MNO coordination and enhance the deployability of the model.

\item Extensive validation is performed using synthetic and real-world Amazon spot price data from diverse regions, demonstrating that the model achieves significant cost reductions while maintaining scalability and robustness under various real-world constraints.

\end{itemize}

The rest of the paper is structured as follows: Section II discusses related papers. Section III introduces the scenario description and problem formulation. Section IV addresses the bandwidth reservation problem in a multi-MNO environment and describes the deep Q-learning model implemented in this work. In Section VI, we present the performance of our proposed model compared with state-of-the-art methods. Finally, a conclusion is drawn in Section VII.

\section{Related Work}
\subsection{Vehicular Resource Reservation}

In recent, vehicle-aided edge computing architecture has gained significant attention from MNOs, such as China Mobile and Vodafone \cite{xiong2019dynamic}. Numerous studies have been devoted to investigating resource provisioning issues within this architecture, as well as in mixed-edge and cloud computing architectures \cite{chen2020edge, zang2019filling,al2022optimal, zang2021soar, gu2016efficient, cao2016share, chen2020stackelberg, liwang2022overbooking, zhang2023dynamic, al2024optimizing, al2024blockchain, al2024resources, liu2018distributed, du2017computation}. However, despite significant efforts in resource provisioning, few of the above works have considered bandwidth provisioning, particularly concerning economic considerations such as diverse MNO pricing strategies. Furthermore, the prevailing research has predominantly focused on onsite resource provisioning, with limited attention given to the inherent overheads associated with such provisioning methods.

Existing studies on resource provisioning and requests can be divided into two categories: i) spot requests, where vehicles request bandwidth resources for immediate use based on current conditions, and ii) forward requests, where vehicles reserve resources for future use. Generally, most studies related to resource reservation and allocation in mobile edge networks tend to emphasize the former spot request mode. For example, in onsite competitions \cite{cao2016share, chen2020stackelberg}, users compete for resources through various game-theoretic methods, such as auctions and Stackelberg games. However, only a limited number of winners acquire the resources, leading to a risk of failure for some users, which may not be suitable for time-critical applications. Onsite requests often exhibit fluctuating pricing and inherent inequities due to the uncertain nature of resource availability and demand.
 
In contrast, immediate requests have been discussed in many challenges, such as bandwidth \cite{niyato2008competitive}, and computing resources in edge networks \cite{chen2020edge, zang2019filling, zang2021soar}. In \cite{niyato2008competitive}, the focus is on sharing the available spectrum between multiple secondary users and a primary user. Here, primary users offer pricing information to secondary users, allowing them to reserve spectrum and optimize their utility. Chen et al. \cite{chen2020edge} developed a meta-learning approach to assist in reserving resources for computing with the goal of minimizing the cost of using edge services. Zang et al. proposed a smart online reservation framework to minimize the cost of reserving resources for an individual user \cite{zang2019filling} or multiple users \cite{zang2021soar}. These approaches typically operate on an immediate request basis, requiring advance reservations due to limited resources. Efficient reservation planning becomes a challenge as users lack knowledge about cost trends and available resources, making it difficult to ensure cost-effectiveness. Additionally, an energy-aware resource request approach under edge computing-assisted UAV networks was investigated \cite{liwang2021let}, where forward or in advance request design and power optimization problems were carefully analyzed.

Similar studies related to forward requests involve computing resource reservations in advance \cite{li2013futures, sheng2019futures}, which are fulfilled accordingly during each future utilization. This guarantees resources without incurring extra delay in decision-making and can thus achieve commendable time efficiency. Li et al. \cite{li2013futures} considered a futures request to manage the financial risk associated with spectrum requests and discover future prices.  In another study \cite{sheng2019futures} Sheng et al. introduced a futures-based spectrum request scheme to achieve mutually beneficial use of resources while alleviating unexpected request failures caused by price fluctuations. Futures-based or advanced request cost-effective resource provisioning in edge network environments has rarely been focused on in existing works. We are among the few to discuss this problem with consideration of economic-related issues with different pricing strategies. In our prior research, a smart place reservation request for time-critical applications was proposed, enabling the advanced reservation of mobility locations at specific time intervals, achieving commendable cost-effectiveness and time efficiency \cite{al2022optimal}. A study by \cite{al2022heuristic} has investigated the issues associated with advanced reservation updates, focusing on minimizing the initial reservation and provisioning costs under uncertainty in demand and price. However, their considered system model investigates the single MNO environment.

\subsection{DRL in Edge Computing for Resource Provisioning}
DRL has emerged as a powerful tool for addressing resource provisioning challenges in edge computing, particularly for vehicle applications. In a study \cite{gazori2020saving}, the authors leverage DRL to optimize the scheduling of IoT applications within a fog computing environment. Effective task scheduling in this context is crucial for minimizing latency, reducing costs, and enhancing overall system performance. In \cite{ning2019deep}, DRL is utilized to optimize the offloading of computation tasks from vehicles to nearby edge computing resources. This approach optimizes resource usage by making dynamic offload decisions based on changing conditions and requirements in the vehicle environment. Furthermore, in \cite{baek2020heterogeneous}, the authors manage the execution of complex tasks in multi-fog networks. They employ DRL to optimize task offloading and resource allocation, considering the limited computing capacity of fog nodes. The objective is to maximize the successful processing of tasks while ensuring compliance with corresponding delay constraints for each task.
Additionally, within the context of reservation systems presented in \cite{zhang2023dynamic}, DeepReserve is introduced as the first dynamic resource reservation system in edge networks utilizing DRL. It dynamically allocates edge servers for connected vehicles, with the primary goal of optimizing the allocation of edge computing resources to enhance the performance of connected vehicle applications. However, their optimization problem does not include monetary cost considerations.

\begin{table*}
    \caption{Notations and their definitions.}
    \begin{tabular} {|c|c|c|c|}
    \hline
    \textbf{Notations} & \textbf{Definition} & \textbf{Notations} & \textbf{Definition} \\
    \hline\hline
    
        $v_h$ & vehicle request bandwidth resource  & $DP_h$ & planned driving path of vehicle $v_h$
        \\ \hline
        
        $RS_i$ & $i^{th}$ road segment & 
        $[\tau_{\text{start}}, \tau_{\text{end}}]$ & time interval of between entering and exiting a particular RS
       \\ \hline
       $BS_i$ & base stations of the $i^{th}$ RS for the $MNO_m$
       &
       $[\tau_k , \tau_w]$ & a specific reference time interval for each cost area A
       \\  \hline
       $\tau_D$ & decision time
       &
        $S_0$, $S_N$ & pricing sessions\\
        \hline
       $\Delta T_h$ & time required for the $v_h$ to pass through the $DP$
       &
        $P_{T_0}$, $P_{T_f}$ & the list of the pricing sessions$S_0$, $S_N$
       \\ \hline
       $ M$ & number of MNOs
       &        
       $St$ & time steps in desired departure interval [$T_0 , T_f$]\\
        \hline
       $ N $ & number of sessions
       &
       $ \left | St\right| $ & number of St
       \\ \hline
       $R$ & number of desired departure time in interval $[T_0 , T_f ]$ 
       &
       $\Delta T_{sv}$  & session validity time\\ 
        \hline
       $p_{n,m,r}$ & price for the $n$ session, $m$ MNO, and $r$ departure time &
        $P(s' \mid s, a)$ & transition probability of state $s'$ from state $s$ and action $a$ 
        \\ \hline
       $\mathcal{S}, \mathcal{A}, \mathcal{R}, \mathcal{P}_r$ & state, action, reward and state transition space    
       &
       $R(s_t, a_t)$ & reward function
       \\ \hline
       $s_t, a_t, R_t$ & state, action and reward at time step $t$
       &
       $V_A(s, a; \theta_a)$ & reward associated with taking action $a$ \\  & & &
       in state $s$ with parameters $\theta_a$
       \\ \hline
       $P_{\text{global min}}^A$ & global minimum price in area $A$ &
       $V(s; \theta_v)$ & expected cumulative reward in state
         $s$ with parameters $\theta_v$ 
       \\ \hline
       $R_{\text{global min}}^A$ & reward for finding global minimum in area $A$ &
       $R_{\text{timeout}}^A$ & penalty for timeout without action in area $A$
    \\ \hline
    \end{tabular}
\end{table*}

While these studies contribute significantly to addressing resource provisioning challenges, this study explores a new aspect of cost-effective resource provisioning through advance requests within a complex environment involving multiple MNOs. To the best of our knowledge, this study is among the first to investigate the cost-effective and efficient bandwidth reservation request problem in an edge network involving various MNOs. This approach aims to mitigate the risks associated with dynamic pricing models of MNOs, providing a comprehensive understanding of cost-effective resource allocation in advance requests.

\section{Scenario Description and Problem Formulation}

\begin{figure}[t]
    
    \hspace{-7pt}
    \includegraphics [width=0.49\textwidth]{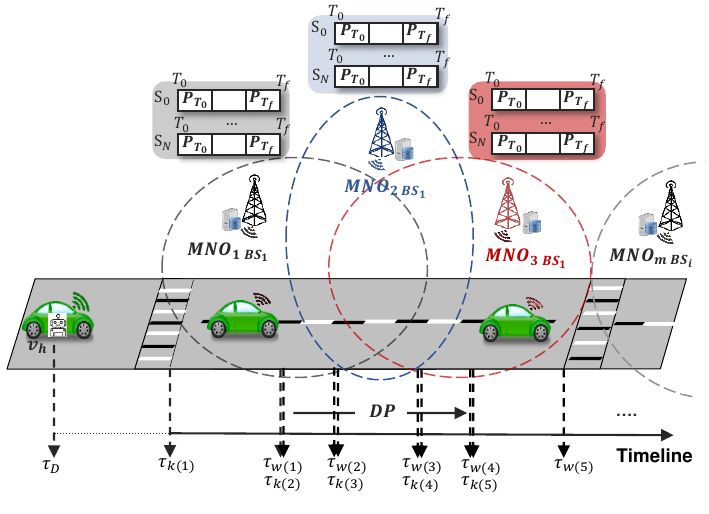}
    
    \caption{An illustration of scenario description}
    \label{fig:system_model}
\end{figure}

In this paper, we model the urban vehicle network as a combination of vehicles and Base Stations (BSs), as illustrated in Fig. \ref{fig:system_model}. BSs can take the form of either roadside units (RSUs) or Macro Base Stations (MBSs). According to their network coverage, the driving path (DP)\footnote{The assumption is that the path refers to the sequence of road segments leading to the destination.} is divided into road segments (RS)\footnote{Road segments are defined as stretches between handoff points and intersections or between two intersections, designated by location pairs $(a, b)$. A handoff point is the location where a road intersects the boundary of a BS coverage area.}, each with a certain number of operating MNOs. These MNOs provide wireless connections between edge routers and connected vehicles within the core network. Within each segment, all MNOs have a dedicated BS at multiple locations along the DP. We refer to the set of all available BSs at road segment $i$, denoted as $MNO_m BS_i$ where $m = 1, 2, \ldots$ and $i = 1, 2, \ldots$.

To meet the stringent latency requirements of vehicular applications, a fog/edge server node is embedded in the infrastructure of each BS. At time $\tau_D$, a vehicle $v_h$ decides to initiate a reservation request to MNOs in advance for each road segment $RS_i$ along the $DP_h$. This time is called decision time. The request specifies the bandwidth time period $\Delta T_h$ needed to complete the intended $DP_h$. The time period $\Delta T_h$ is divided into reservation time intervals $[\tau_{\text{start}}, \tau_{\text{end}}]$ for each $RS_i$ or each coverage areas\footnote[3] {The diameter of a $BS$ coverage area is approximately 900m, similar to \cite{soh2006predictive}, and it is typically visible in downtown areas of metropolises.} of $MNO_m BS_i$, where $\tau_{\text{start}}$ is the entry time into a $RS_i$, and $\tau_{\text{end}}$ is the exit time from that segment.
These intervals depend on the length of $DP_h$ and the speed of the vehicle, which are calculated based on information from the navigation system. In addition, vehicle $v_h$ divides its desired departure interval $[T_0 , T_f]$ into $\left | St\right|$ small intervals (i.e., time steps). The vehicle $v_h$ requests the cost for its route $DP_h$ for each possible $St$ interval. The list of received costs ($P_{T_0}, P_{T_1}, \ldots, P_{T_f}$) is referred to as session $S_0$. It repeats this procedure every $\Delta T_{sv}$, $N$ times maximum, and accordingly receives the pricing sessions $S_0, S_1, \ldots, S_N$ (as shown in Fig. \ref{fig:system_model}). $\Delta T_{sv}$ refers to the pricing sessions that are still valid or the session validity time, which is a parameter determined by MNO.

The multi-MNO bandwidth reservation problem is formulated as a mathematical optimization problem. The goal is to minimize the overall cost of reserving bandwidth over a time interval $[\tau_{\text{start}}, \tau_{\text{end}}]$ for the $DP$. The $DP$ is divided into cost areas where the $MNO_m BS_i$ is located, using a function $\varphi$. Each cost area $A$ corresponds to a specific reference time interval $[\tau_k - \tau_w] \subset [\tau_{\text{start}}, \tau_{\text{end}}]$, $\ k,w\in\mathbb{N}$.

\begin{equation}
    \varphi:\mathbb{R}_+^{2}\rightarrow\mathbb{R}_+^{N\times M\times R}: (\tau_k,\tau_w)\mapsto A
    \label{eq:state}
\end{equation}

where $A$ is a tensor holding the information about the prices:

\begin{equation}
A =
\begin{bmatrix}
\begin{pmatrix}
p_{111} & \cdots & p_{1M1} \\
\vdots & \ddots & \vdots \\
p_{N11} & \cdots & p_{NM1}
\end{pmatrix}
, &

\begin{pmatrix}
p_{112} & \cdots & p_{1M2} \\
\vdots & \ddots & \vdots \\
p_{N12} & \cdots & p_{NM2}
\end{pmatrix}
,.  
..
, 
\\

\begin{pmatrix}
p_{11R} & \cdots & p_{1MR} \\
\vdots & \ddots & \vdots \\
p_{N1R} & \cdots & p_{NMR}
\end{pmatrix}
\end{bmatrix}
\end{equation}

This tensor consists of $R$ slices (representing desired departure time intervals), each containing a matrix with $N$ rows (representing sessions [prices at time of request]) and $M$ columns (representing MNOs), where $p_{n,m,r}$ denotes the price corresponding to the $n$-th session, the $m$-th MNO, and the $r$-th desired departure time.
This representation of the state preserves, the Markov property, so that the agent can take actions based only on the provided state without any need for the previous states history.

In this context, the set $\mathcal{A}_{disjoint}$ encompasses tensors $(A_i)_{i\in I}$ that correspond to segments with the presence of only a single MNO:

\begin{equation}
    \mathcal{A}_{disjoint}:=\{\ (A_i)_{i\in I}\ |\ \# MNO=1\ \}
    \label{eq:state}
\end{equation}

while $\mathcal{A}_{overlap}$  refers to the set of tensors $(A_j)_{j\in I}$ associated with segments that involve the presence of multiple MNOs:
 
\begin{equation}
    \mathcal{A}_{overlap}:=\{\ (A_j)_{j\in I}\ |\ \# MNO>1\ \}
    \label{eq:state}
\end{equation}

Hence, by considering both $(A_j)_{j\in I}$ and $(A_j)_{j\in I}$, the resulting collection is denoted as $\mathcal{C}$ and is structured as follows:

\begin{equation}
    \therefore \mathcal{C}= \mathcal{A}_{disjoint}\cup \mathcal{A}_{overlap}
    \label{eq:state}
\end{equation}

The $\theta$ function is employed to associate a specific Area $A$ with its overall reservation cost:

\begin{equation}
    \theta:\mathbb{R}_+^{N\times M\times R}\rightarrow \mathbb{R}_+
    \label{eq:state}
\end{equation}

while the $\omega$ function, defined as $\omega=\theta\circ\varphi$, maps:

\begin{equation}
    \therefore \omega:\mathbb{R}_+^2\rightarrow\mathbb{R}_+
    \label{eq:state}
\end{equation}

Finally, the objective function $J$ is defined as follows:

\begin{equation}
    J(\tau_{\text{start}}, \tau_{\text{end}})=\sum_{\tau_k, \tau_w \in [\tau_{\text{start}}, \tau_{\text{end}}]} {\omega(\tau_k, \tau_w)}
    \label{eq:state}
\end{equation}

The sum operates over all pairs $(\tau_k, \tau_w)$ of start and end times in the reference interval $[\tau_{\text{start}}, \tau_{\text{end}}]$.
    
The function $\omega$ maps these time intervals to a total reservation cost as per the previously defined composite function (which uses both $\varphi$ and $\theta$ ). To make the problem more practical and realistic, the following constraints are imposed:

\paragraph{Time Interval Constraints}
The reservation period must respect the operational limits of the multi-MNO system:
\begin{equation}
    \tau_{\text{start}} \geq T_{\text{min}}, \quad \tau_{\text{end}} \leq T_{\text{max}}, \quad \tau_{\text{end}} > \tau_{\text{start}}
\end{equation}
where $T_{\text{min}}$ and $T_{\text{max}}$ define the earliest and latest allowable reservation times.

\paragraph{Bandwidth Availability Constraints}
The reserved bandwidth must not exceed the available capacity of each $MNO_m BS_i$ during the specified time interval:
\begin{equation}
    \sum_{n=1}^N \sum_{m=1}^M b_{n,m}(\tau_k, \tau_w) \leq C_{m,i}, \quad \forall \tau_k, \tau_w
\end{equation}
where $b_{n,m}$ represents the reserved bandwidth for session $n$ with MNO $m$, and $C_{m,i}$ is the total bandwidth capacity of the $m$-th MNO’s $i$-th base station.

\paragraph{Cost Area Constraints}
Each reservation must correspond to a valid cost area $A$, ensuring that the tensor mappings through $\varphi$ are properly defined:
\begin{equation}
    A \in \mathcal{C}, \quad \mathcal{C} = \mathcal{A}_{\text{disjoint}} \cup \mathcal{A}_{\text{overlap}}
\end{equation}
This ensures compatibility with scenarios involving overlapping or disjoint MNO coverage.

\paragraph{Reservation Cost Constraint}
To prevent excessive cost, an upper bound on the total reservation cost can be imposed:
\begin{equation}
    J(\tau_{\text{start}}, \tau_{\text{end}}) \leq J_{\text{max}}
\end{equation}
where $J_{\text{max}}$ is a predefined budget limit.

\section{Bandwidth Reservation in Multi-MNO Environment}
\label{MDP}
In this section, we introduce proposed Area-wise approach. Following that, the DRL problem is formally formulated. Subsequently, variants of Q-learning algorithms, such as Deep Q-Network (DQN), Double Deep Q-Network (Double DQN), and Dueling Deep Q-Network (Dueling DQN), are employed in this paper. Additionally, to mitigate the costs associated with direct agent-environment interactions, we also utilize the Temporal Fusion Transformer (TFT) \cite{lim2021temporal} to provide the RL-model with synthetic prices during training. This approach creates a sophisticated training environment that closely resembles real-world pricing dynamics, as illustrated in Fig. 2.

\subsection{Area-wise Approach for Efficient Q-Learning Training}

In this subsection, the strategy of optimizes the Q-learning process by presenting individual areas to the agent during episodes, leading to improved computational efficiency and adaptability.
Since each cost area $A$ is a collection of price samples of different MNOs corresponding to different probability distributions, our goal is to bring the agent to learn these various distribution differences. Consequently, each area $A$ is presented to the RL-agent during each episode. This practice offers a reliable approximation for price determination in what we define as the \textbf{Area-wise approach}.
The approach key components and the benefits it offers are outlined as follows: 

\textit{Single Area Episodes.} In each episode, a single area $A$ is presented to the Q-learning agent from the collection $\mathcal{C}$. This action effectively reduces the state space for the episode, concentrating solely on the price dynamics of that particular area.\\
\textit{Temporal Learning.} Since the agent is dealing with one area at a time, it can learn the temporal dynamics more effectively, understanding the price variations over time, and making decisions based on shorter-term patterns.\\
\textit{Transfer Learning.} If certain areas have similar characteristics (like the same MNOs or similar pricing structures), the knowledge gained from one area can be transferred to another, speeding up the learning process.\\
\textit{Aggregation of Policies.} Once the agent has been trained on individual areas,  the learned policies are aggregated to form a global strategy. The global strategy will consist of optimal actions (bookings) for each area, which can be sequentially executed as the agent traverses through the areas.\\
\textit{Replay Buffer.} a replay buffer is employed in the Q-learning approach. Store experiences from various areas in this buffer. Even if the agent is currently learning about a particular area in an episode, it can sample past experiences from other areas to stabilize and generalize its learning. The Area-wise approach offers several advantages:\\
\textit{Computational Efficiency.} By focusing on one area at a time, the state space for each episode is drastically reduced, making computations faster.\\
\textit{Modularity.} This approach allows for modular learning. If a new area gets added or an existing area undergoes significant changes (like a new MNO entering the market), the agent can be retrained just for that area without disrupting the entire model.\\
\textit{Adaptive Learning.} As pricing dynamics or MNO strategies evolve, the agent can quickly adapt its strategy for affected areas without a complete retraining.

\subsection{Formulation of DRL Problem}
Applying the Area-wise approach will simplify the problem during training to be the following cast of  a MDP, defined by $\mathcal{M} = (\mathbf{S}, \mathbf{A}, \mathbf{P}, \mathcal{R})$:
\begin{itemize}
    \item $\mathbf{S}$ is the state space, encapsulating the varied prices offered by different MNOs at each time instance.
    \item $\mathbf{A}$ is the action space, providing the choice to accept an offer or wait for subsequent propositions.
    \item $\mathbf{P}$ stands for the transition probability kernel, a pivotal component capturing how states evolve over time based on certain actions. 
    \item $\mathcal{R}$ is the reward function, offering a mathematical quantification of the immediate benefit or disadvantage of every action.
\end{itemize}

The design of the state space, action space, and reward in the DRL model is as follows:

\subsubsection{State Space}
For each area $A$ of time segment $[\tau_k, \tau_w]$, the state space is defined as the probability distribution of prices offered by different MNOs. These distributions can be represented using a probability density function (PDF):

\begin{equation}
    s_t^A = [p_{1,m,r}, p_{2,m,r}, p_{3,m,r}, \cdots, p_{N,m,r}]
    \label{eq:state}
\end{equation}

Where $p_{n,m,r}$ is  the price at the $n$-th session offered by the $m$-th MNO for the $r$-th desired departure time during time $t$ of making decision.

Since we need an initial state $S^A_0$ for training the RL-agent, we can take any existing environment with states $S^A_1,S^A_2,S^A_3...S^A_n$ plus defined actions and rewards. Add a special fixed start state $S_0$ with one action $\alpha^A_{wait}$ (wait for next offer) and reward $0$. It is clearly a valid MDP, and is identical to the original MDP in terms of value and policy functions for states $S^A_1,S^A_2,S^A_3...S^A_n$. For all intents and purposes to the agent (which gets no meaningful policy choice in $S^A_0$), it starts in the original MDP in some random state.
\subsubsection{Action Space}

The action space for each area $A$ and time segment $[\tau_k, \tau_w]$ is:
\begin{equation}
    \mathbf{A}^A_t = \{\alpha_{n,m,r}\} \cup \{\alpha_{\text{wait}}^A\}
    \label{eq:state}
\end{equation}

Where $\alpha_{n,m,r}$ corresponds to selecting the offer from  $MNO_m$ at the $n$-th session for the $r$-th desired departure time during time $t$ of taking decision in area $A$ and $\alpha_{\text{wait}}^A$ symbolizes waiting for future offers.

\subsubsection{Objective Function}

Given the stochastic nature of prices from MNOs, the objective function for each area $A$ within the time segment $[\tau_k, \tau_w]$ aims to minimize the expected reservation cost:

\begin{equation}
    J^A = \mathbb{E}\left[\omega^A(\tau_k,\tau_w)\right]
    \label{eq:state}
\end{equation}

Here, $\mathbb{E}[\cdot]$ reflects the stochastic nature of prices from MNOs as encapsulated by the state space. The optimization task is to find a policy $\pi^*_t$ minimizing the expected cumulative cost over all areas:

\begin{equation}
    \pi^* = \arg\min_\pi \mathbb{E}\left[\sum_{A} \mathcal{C}(\mathbf{A}^A, \mathbf{S}^A) | \pi\right]
    \label{eq:state}
\end{equation}

\subsubsection{Reward Signal:}

The reward signal $\mathcal{R}^A(\mathbf{A}^A_t, \mathbf{S}^A_t)$ quantifies immediate outcomes:

\begin{align}
    \mathcal{R}^A(\mathbf{A}^A_t, \mathbf{S}^A_t) =
    \begin{cases}
    R_{\text{global min}}^A & \text{for finding global} \\
    & \text{minimum}, \\
    -e^{-h \cdot t} & \text{for waiting}, 
    \\
    -R_{\text{timeout}}^A
        & \text{penalty for timeout} \\
        & \text{without action}
    \\
    p_{\text{global min}}^A - p_i(t) & \text{otherwise}
    \end{cases}
    \label{eq:state}
\end{align}

\begin{itemize}
    \item Selecting the globally minimal price yields a reward $R_{\text{global min}}^A$, signifying a positive benefit.
    \item Choosing $p_i(t)$ that isn't the lowest results in a reward of $p_{\text{global min}}^A - p_i(t)$, representing the opportunity gain.
    \item Waiting until time $t$ incurs a penalty of  $-e^{-h \cdot t}$, where $h$ is a hyperparameter, adjusted during training, discouraging delays.
    \item If the decision time elapses without any action, the agent receives a reward $-R_{\text{timeout}}^A$ penalty.
\end{itemize}

\subsection{Design of Framework}
\begin{figure}[t]
    \centering
    \includegraphics[width=8.5cm, height=8cm]{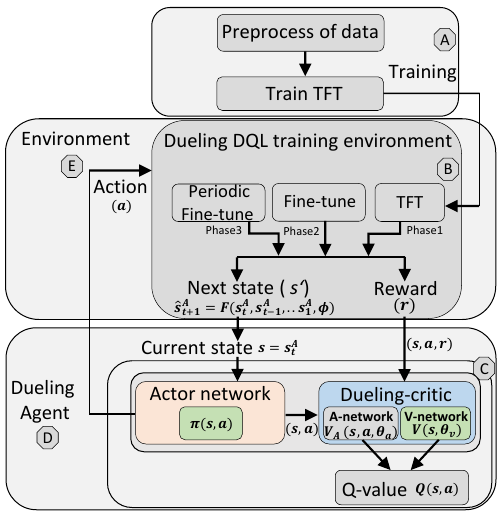}
    \caption{An illustration of the proposed framework for bandwidth reservation with multi-phase training.}
    \label{fig:framework}
\end{figure}

In this section, the framework consists of two components, as depicted in Fig. \ref{fig:framework}, mirroring the conventional RL paradigm found in the traditional loop diagram. The traditional loop diagram in DRL involves a cyclical process where an agent interacts with an environment. The agent observes the current state of the environment, takes action based on its policy, and receives a reward from the environment. This feedback loop is essential for the agent to learn and update its policy, aiming to maximize cumulative rewards over time. The loop repeats iteratively, allowing the agent to refine its decision-making through continuous interaction with the environment. Fig. \ref{fig:framework}E corresponds to the environment in the traditional loop diagram, while Fig. \ref{fig:framework}D corresponds to the agent.

In the initial step, historical data (Fig. \ref{fig:framework}A) is utilized to construct TFT model (Fig. \ref{fig:framework}A), addressing previously mentioned issues. The choice of the TFT method is rooted in its capacity to achieve higher accuracy in time-series predictions. The preprocessing of raw data and the training of TFT models (\ref{fig:framework}A) are one-time procedures that yield inputs for establishing the DRL training environment in the subsequent step (Fig. \ref{fig:framework}B).

The DRL training environment (Fig. \ref{fig:framework}E) receives an action ($\alpha_{wait}^A$), any other action will result in ending the training episode, because it results in accepting the prices offered. In (Fig. \ref{fig:framework}B) depending on the training phase the state is either send from the TFT if it's phase 1 otherwise it is phase 2 and the state $s^\prime$ is extracted from the real data table used previously for training the TFT model, the reward $r$ is respectively calculated. The interaction between the DRL agent and the DRL training environment is iterated until the DRL agent converges to an optimal policy, forming the second part of the framework.

In Fig. \ref{fig:framework}D, The Dueling DQN architecture is composed of two main components: the dueling actor network and the dueling critic network. In Fig. \ref{fig:framework}C, the actor network is illustrated, taking the current state $s$ as input and selecting an action $a$. Simultaneously, the dueling critic network takes the current state $s$, action $a$, and reward $r$ as inputs, generating the quality of the action $Q(s,a)$. This quality represents the anticipated cumulative reward the agent expects to receive.

In contrast to a traditional DQN, the dueling architecture separates the estimation of the state value and the advantage of each action. The dueling critic network computes two values: the state value, denoted as $V(s)$, which captures the value of being in the current state irrespective of the chosen action, and the advantage value, denoted as $A(s, a)$, which represents the additional value associated with taking a specific action in the given state.
The dueling actor network is responsible for combining these two values, utilizing a mechanism that facilitates the simultaneous estimation of both the state value and the advantages of each action. This separation enhances the learning process by allowing the agent to focus on states where either the state value or the advantage value are particularly informative.
During training, the Dueling DQN agent aims to minimize the cost of reservation by iteratively selecting actions that lead to the lowest price and the optimal time of departure. The learning process involves observing the quality of its actions over a training period. This iterative training continues until the agent's reward converges to a stable value, resulting in the development of an optimal policy.
Detailed explanations of each component of the dueling DQN framework are provided in subsequent sections, elucidating how the dueling architecture enhances the learning and decision-making capabilities of the agent.

\subsection{Suitability of Q-Learning}

In the context of the area-wise approach, it becomes evident that the decision-making process of accepting a price or waiting for a potentially better one resembles the well-known \textit{secretary problem} \cite{fujii2023secretary}. This problem involves making decisions promptly upon observing each candidate (or price, in our case). The primary challenge lies in identifying the optimal moment to make a commitment. 
Since the \textit{secretary problem} can be represented through the Bellman equation.
Our problem can then be an applied form of 
Q-learning. For our problem, the Bellman equation is:

\begin{equation}
    \ Q(s, a) = \mathcal{R}(a, s) + \gamma \sum_{s' \in \mathcal{S}} \mathcal{P}(s' | s, a) \max_{a' \in \mathcal{A}} Q(s', a') \
    \label{eq:state}
\end{equation}

The action taken in this particular scenario depends on the current state and has future consequences, a characteristic central to the Bellman equation. The primary goal in both problems is to optimize a decision based on current and potential future rewards.

Q-learning iteratively updates Q-values using the temporal difference (TD) method:

\begin{equation}
    \Delta Q(s, a) = \alpha \left( \mathcal{R}(a, s) + \gamma \max_{a' \in \mathcal{A}} Q(s', a') - Q(s, a) \right)
    \label{eq:state}
\end{equation}

By utilizing deep Q-networks (DQN), the Q-function can be approximated using neural networks, allowing for efficient learning in high-dimensional spaces like this bandwidth reservation problem. In Deep Q-Learning, a deep neural network is employed to approximate the various Q-values for each potential action within a state (value-function estimation). The distinction lies in the training phase, where, rather than directly updating the Q-value of a state-action pair as traditionally done in Q-Learning, a loss function is formulated. This loss function assesses the disparity between our Q-value prediction and the Q-target and gradient descent is applied to adjust the weights of the Deep Q-Network for more accurate Q-value approximation. Then, the loss function $L(\theta) $ for the network with parameters $\theta $ is defined as follows:

\begin{equation}
    L(\theta) = \mathbb{E}\left[ \left( \mathcal{R}(a, s) + \gamma \max_{a' \in \mathcal{A}} Q(s', a'; \theta^-) - Q(s, a; \theta) \right)^2 \right]
    \label{eq:state}
\end{equation}
where $\theta^-$ are the parameters of a target network.

For Double Deep Q-Network (DDQN), the loss function changes to:

\begin{align}
    L(\theta) &= \mathbb{E}\left[ \left( \mathcal{R}(a, s) + \gamma Q(s', \arg\max_{a' \in \mathcal{A}} Q(s', a'; \theta); \theta^-) \right) \right. \notag \\
    &\quad - Q(s, a; \theta) \big)^2 \bigg]
    \label{eq:state}
\end{align}

For Dueling DQN:

Given:

\begin{equation}
    Q(s, a; \theta) = V(s; \theta_v) + V_A(s, a; \theta_a)
    \label{eq:state}
\end{equation}
Where $\theta_v$ and $\theta_a$ are parameters for the value and advantage functions, respectively. The loss function can be similarly defined with the above expressions.

\subsection{Incorporation of Temporal Fusion Transformer (TFT):}

Given the state $s_t^A$, TFT helps us estimate future states $\hat{s}_{t+1}^A, \hat{s}_{t+2}^A,\cdots$. Mathematically, if $F$ is our TFT, then:

\begin{equation}
    \hat{s}_{t+i}^A = F(s_t^A, s_{t-1}^A, ..., s_1^A; \phi)
    \label{eq:state}
\end{equation}
Where $\phi$ are the parameters of TFT.

By integrating TFT with Q-learning, we obtain a richer estimate of the Q-values:

\begin{align}
Q&(s^A, a^A; \theta, \phi) \approx \mathcal{R}(a^A, s^A) \notag \\
&+ \gamma \sum_{i=1}^N \mathcal{P}(\hat{s}_{t+i}^A | s^A, a^A) \max_{a'^A \in \mathbf{A}} Q(\hat{s}_{t+i}^A, a'^A; \theta)
\label{eq:state}
\end{align}

This enhanced algorithm provides a more detailed walkthrough of the Dueling DQN, emphasizing the distinct value and advantage streams and how they are incorporated into the Q-learning process.

\subsection{Multi-Phase Training Algorithm}
The presented algorithm, named "Enhanced Dueling DQN for Bandwidth Reservation with Multi-Phase Training," is a novel approach designed for addressing bandwidth reservation challenges in dynamic environments. The algorithm integrates two key components: Dueling DQN and TFT. The objective is to optimize the decision-making process in a dynamic setting, where both synthetic and real-world data contribute to the training process.

The algorithm begins by initializing the state (\(S_0\)), the Dueling Q-network (\(Q\)) with separate streams for state value (\(V\)) and advantage value (\(V_A\)), experience replay memory (\(D\)) with capacity (\(C\)), discount factor (\(\gamma\)), and exploration factor (\(\epsilon\)). The replay memory \(D\) and the Q-network \(Q\) are initialized with random weights (\(\theta\)), and a target network (\(Q'\)) is created as a copy of \(Q\)
\textit{(line 1-3 in the algorithm)}.
\\
\textbf{Phase 1: Synthetic Training on TFT-generated Data}

In the first phase, the algorithm undergoes synthetic training on data generated by the TFT. It iterates through episodes (\(E_{\text{synthetic}}\)), where the agent interacts with the environment by requesting a session from available MNOs in the area $A$ and then selects actions based on an \( \epsilon \)-greedy policy derived from the Q-network. The interactions are performed in a loop until the state becomes terminal. During each step, the algorithm stores the experienced transition in the replay memory \(D\) and updates the Q-network using mini-batches sampled from \(D\).

The key innovation in this phase lies in the decomposition of the Q-value into state value (\(V\)) and advantage value (\(V_A\)). The target value (\(y\)) is computed, considering both the immediate reward and the temporal difference with respect to the forecasted prices from TFT. The Q-network weights (\(\theta\)) are updated through gradient descent on the temporal difference error
\textit{(line 4-18 in the algorithm)}.

\textbf{Phase 2: Dataset Training on Data that Trained TFT (fine-tuning)}

In the second phase, the algorithm transitions to training on a real dataset that initially trained the TFT. The training procedures in this phase resemble those in Phase 1, but the data is drawn from the real-world training set used for TFT. This phase aims to fine-tune the agent based on actual historical data, bridging the gap between synthetic and real-world scenarios
\textit{(line 19-21 in the algorithm)}.

\textbf{Phase 3: Periodic Online Fine-Tuning on Accumulated Environmental Data using Dueling DQN}

In the third and final phase, the algorithm engages in periodic online fine-tuning on accumulated environmental data. While the environment is active, the agent continuously acquires new data $(S, A, r, S')$ from interactions with the environment and stores these transitions in the replay memory $D$. To ensure that the agent adapts to changing environmental conditions, fine-tuning is performed at regular intervals determined by the fine-tuning timer ($T_{\text{fine-tune}}$).

When the timer exceeds the predefined fine-tuning interval, the algorithm samples a mini-batch $B$ of transitions from the replay memory $D$. Similar to the previous phases, the Q-value is decomposed into state value $V$ and advantage value $V_A$, and the target value $y$ is computed using the temporal difference with respect to the updated forecasted prices from TFT. 

\begin{algorithm}[H]
\caption{Enhanced Dueling DQN for Bandwidth Reservation with Multi-Phase Training}
\begin{algorithmic}[1]

\REQUIRE Initial state \( S_0 \), Dueling Q-network \( Q \) with separate streams for \( V \) and \( V_A \), Experience replay \( D \) with capacity \( C \), Discount factor \( \gamma \), Exploration factor \( \epsilon \), Fine-tuning interval \( T_{\text{fine-tune}} \)

\STATE Initialize replay memory \( D \) of capacity \( C \)
\STATE Initialize Q-network \( Q \) with random weights \( \theta \)
\STATE Initialize target network \( Q' \) with weights \( \theta' = \theta \)
\STATE Set a timer for fine-tuning interval \( T_{\text{fine-tune}} \)

\COMMENT{Phase 1: Synthetic Training on TFT-generated data}
\FOR{episode \( e = 1 \) to \( E_{\text{synthetic}} \)}
    \STATE Initialize the state \( S \)
    \WHILE{\( S \) is not terminal}
        \STATE Choose action \( A \) using an \( \epsilon \)-greedy policy derived from \( Q \)
        \STATE Execute action \( A \), observe reward \( r \) using \( \mathcal{R}(A_t, S_t) \), and next state \( S' \)
        \STATE Store transition \( (S, A, r, S') \) in \( D \)
        \STATE Sample random mini-batch \( B \) of transitions from \( D \)
        \FOR{each transition in \( B \)}
            \STATE Decompose \( Q(S, A; \theta) \) into value \( V(S; \theta, \alpha) \) and advantage \( V_A(S, A; \theta, \beta) \)
            \STATE Compute \( y \) as \( r \) if terminal, else \( y = r + \gamma V(S'; \theta', \alpha') + (V_A(S', A; \theta', \beta') - \text{avg}_{a'} V_A(S', a'; \theta', \beta')) \)
            \STATE Update \( \theta \) via gradient descent on \( \left( y - Q(S, A; \theta) \right)^2 \)
        \ENDFOR
        \STATE Update \( \theta' \) with \( \theta \) every \( C \) steps
    \ENDWHILE
\ENDFOR

\COMMENT{Phase 2: Dataset Training on data that trained the TFT}
\FOR{episode \( e = 1 \) to \( E_{\text{dataset}} \)}
    \STATE Procedures similar to Phase 1 but use real dataset that trained the TFT
\ENDFOR

\COMMENT{Phase 3: Periodic Online Fine-Tuning on Accumulated Environmental Data}
\WHILE{environment is active}
    \STATE Acquire new data from the environment: \( (S, A, r, S') \)
    \STATE Store the new transition \( (S, A, r, S') \) in \( D \)
    \IF{timer exceeds \( T_{\text{fine-tune}} \)}
        \STATE Sample a mini-batch \( B \) of transitions from \( D \)
        \FOR{each transition in \( B \)}
            \STATE Decompose \( Q(S, A; \theta) \) into value \( V(S; \theta, \alpha) \) and advantage \( V_A(S, A; \theta, \beta) \)
            \STATE Compute \( y \) as \( r \) if terminal, else \( y = r + \gamma V(S'; \theta', \alpha') + (V_A(S', A; \theta', \beta') - \text{avg}_{a'} V_A(S', a'; \theta', \beta')) \)
            \STATE Update \( \theta \) via gradient descent on \( \left( y - Q(S, A; \theta) \right)^2 \)
        \ENDFOR
        \STATE Update \( \theta' \) with \( \theta \) every \( C \) steps
        \STATE Reset the fine-tuning timer
    \ENDIF
\ENDWHILE 
\end{algorithmic}
\end{algorithm}

The Q-network weights $\theta$ are then updated through gradient descent on the temporal difference error. 
Additionally, the target network $(Q')$ is updated by synchronizing its weights with the current Q-network $(Q)$ every $C$ steps. This helps stabilize the training process by providing more consistent target values during the updates. The fine-tuning timer is reset after each round of fine-tuning to initiate the countdown for the next fine-tuning interval. This phase ensures that the agent remains adaptive to evolving environmental dynamics throughout its deployment
\textit{(line 22-33 in the algorithm)}.
\section{PERFORMANCE EVALUATION}

In this section, we present the results of detailed experimental studies. To demonstrate the efficiency of proposed method, we compare our algorithm with various existing bandwidth reservation schemes in vehicular settings.

\subsection{Experimental Settings}

Dataset Description. To assess the effectiveness of our methodology, we utilized a historical dataset of Amazon spot prices, which are subject to fluctuations influenced by factors such as capacity, demand, geographic location, and specific instance types \cite{amazonspot}. Given the time-sensitive nature of various applications, vehicles require both computing instances and communication links, i.e., bandwidth. Our assumptions, the pricing for setting up computing and communication resources aligns with Amazon's spot pricing model, as previously referenced in \cite{zang2021soar, zang2019filling}. For this study, we collected pricing data from all available instances and two specific regions, namely us-west-1b and us-west-1c. This data was collected from April 17 / 2021 to May 2 / 2021 for training purposes, and from May 3 / 2021 to May 8 / 2021 for the testing phase of the model. In addition, we utilized more recent data, referred to as real-data, from August 17 to October 21 of year (2023) to test the performance of the models in phases 1 and 2 of the algorithm. The evaluation was conducted using the average cost savings metric (see Fig. 9, Fig. 10, and Fig. 11).

System Setup. In the system, the total number of desired departure interval [$T_0-T_f$] is randomly drawn from [8-32]. As a result, each training episode will consist of a randomly selected session group, with a group size ranging from a minimum of 8 sessions to a maximum of 32 sessions. A random session group represents a price area $A\subset\mathcal{C}$, where $\mathcal{C}$ is the collection of price areas discussed in the problem formulation. The time between each two sessions $T_{sv}$ is set to $1$ minute in the experiment.   
A reward function, influenced by the hyperparameter set to 0.01, serves as a cost computed in each episode. The total number of episodes is set at 1000. The positivity of this hyperparameter is positive because when the waiting time increases, the penalty becomes bigger in absolute value. To explain, if the waiting is 10 minutes the reward is -1.105171, while the waiting is 20 minutes, the reward is -1.221403.
Meanwhile, the discount factor $\gamma$ for the Q-learning agents was set to $\gamma=0.85$, prioritizing future offers and encourages exploration, while the exploration factor $\epsilon$ was set to $\epsilon=0.05$, meaning that with a probability of 5\% the agent will choose a random action. The 5\% probability ensures periodic exploration of new actions, mitigating the risk of the agent getting stuck in suboptimal strategies and enhancing its adaptability in diverse environments. The initial session $S^A_0$ is simply a session with zero prices, and this is for every price area, so $S_0=S^A_0$. This meticulous configuration of phases and parameters ensures the development of a robust and adaptive system capable of effectively navigating complex real-world environments. In our experimental setup, we have tested the model in over 20 road segments, each containing three MNOs, to evaluate its performance under varying traffic and network conditions.

Training Configuration: The model is constructed using the PyTorch library (version 2.0.0). All experiments have been conducted on a Windows workstation with the following specifications: CPU - Intel(R) Core(TM) i7-8750H CPU @ 2.20GHz, RAM - 16.0 GB DDR4, GPU - NVIDIA GeForce GTX 1050 Ti with Max-Q Design. The computational requirements of the model are 101.38 KMAC and 202.75 KByte VRAM. When compared to the NVIDIA AGX Xavier, which is recognized as the most powerful System-on-Chip (SoC) \cite{liu2020computing}, the algorithm utilization is only $4*10^{-5}$\% of the on-board computation power of an autonomous vehicle, which can be considered negligible.

Metrics. Four metrics are utilized to evaluate our model: 1) average cumulative reward, i.e., average reward during training episodes for an agent trained; 2) cumulative cost, which is the profit gained from applying our model; 3) penalty costs, which represents the repercussions faced by the agent due to waiting time; 4) an examination of the relationship between decision time and reservation time in our model.

\begin{figure*}[t]
    \centering
    \hspace{-50pt}
    \subfloat[LSTM]{\includegraphics[width=0.38\textwidth]{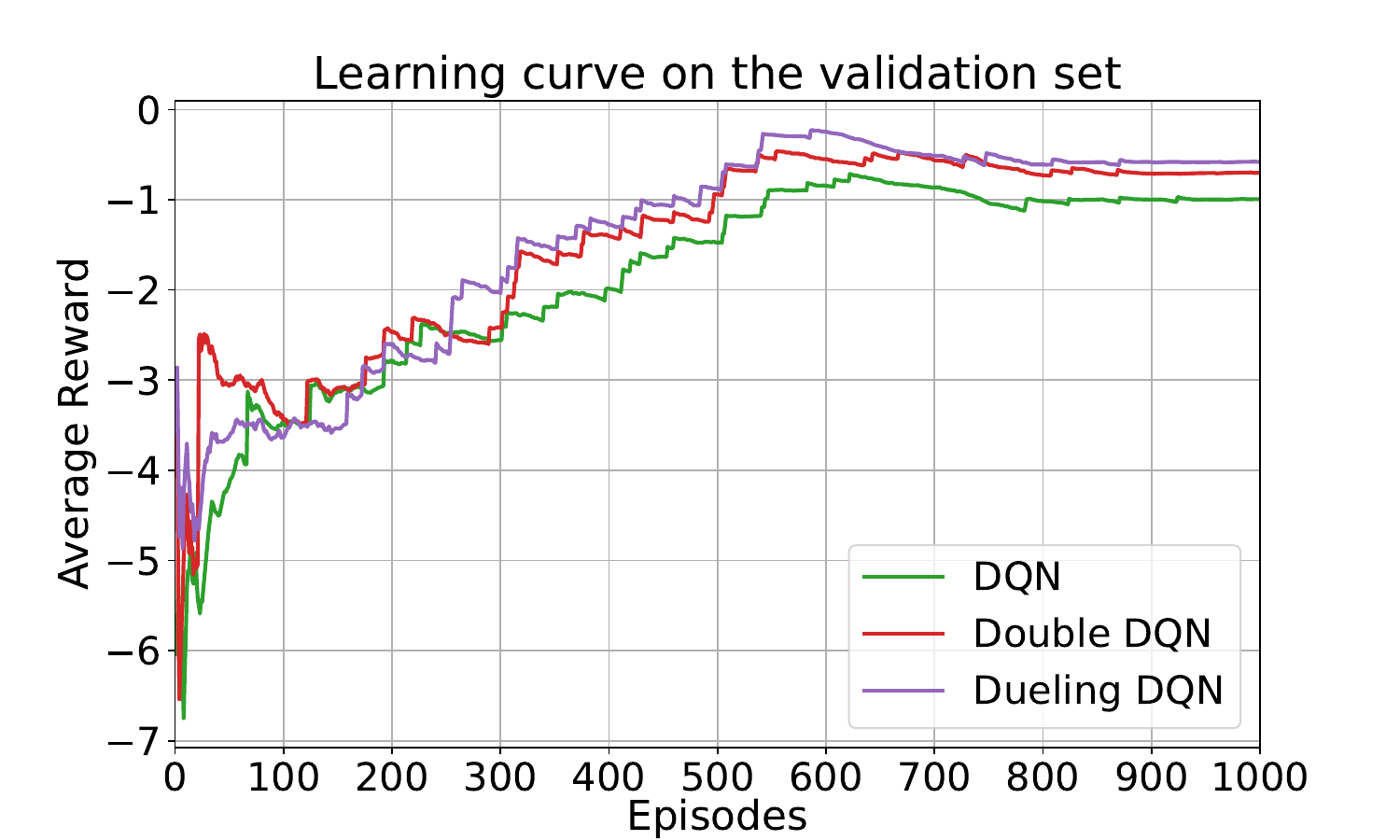}}
    \hspace{-16pt}
    \subfloat[Transformers]{\includegraphics[width=0.38\textwidth]{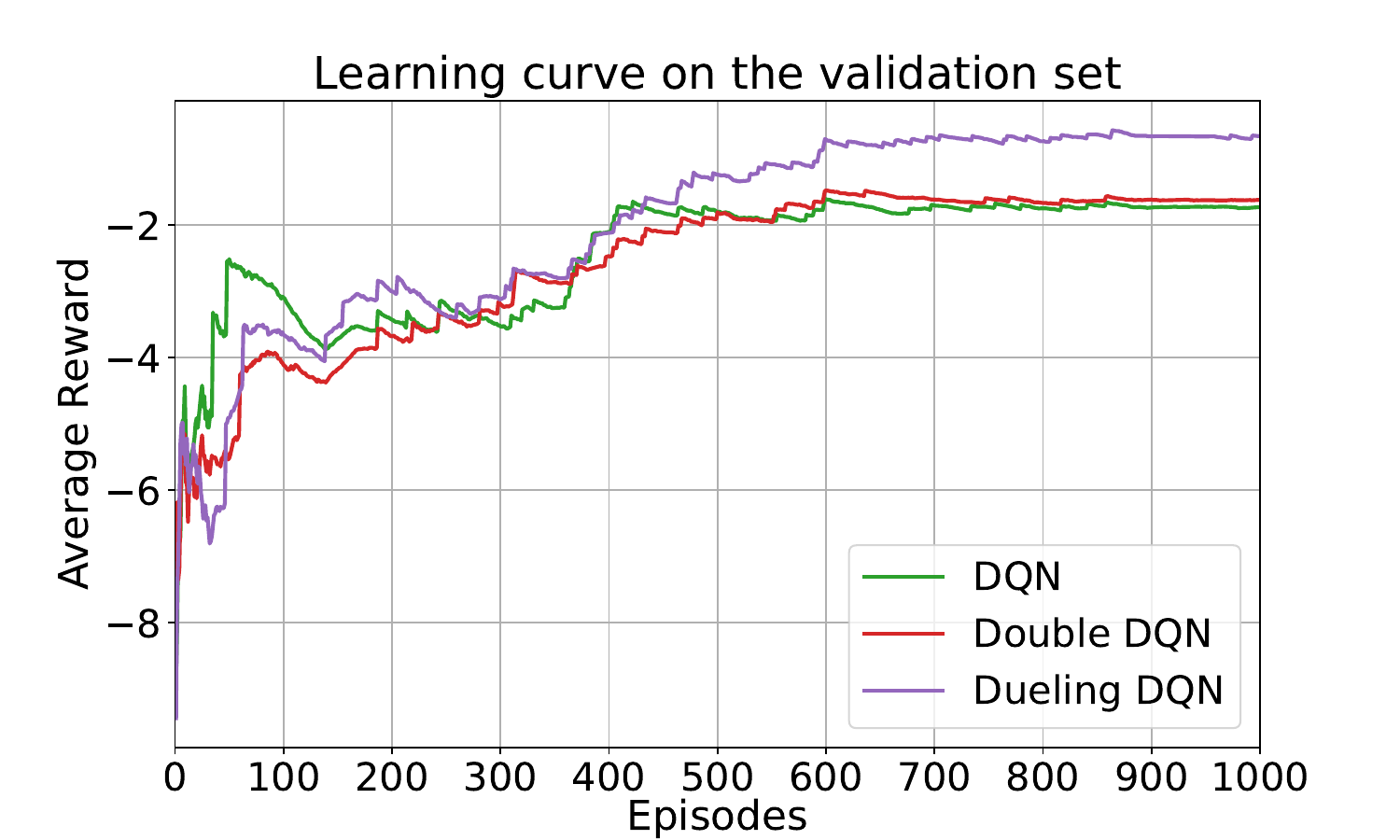}}
    \hspace{-16pt}
    \subfloat[TFT]{\includegraphics[width=0.38\textwidth]{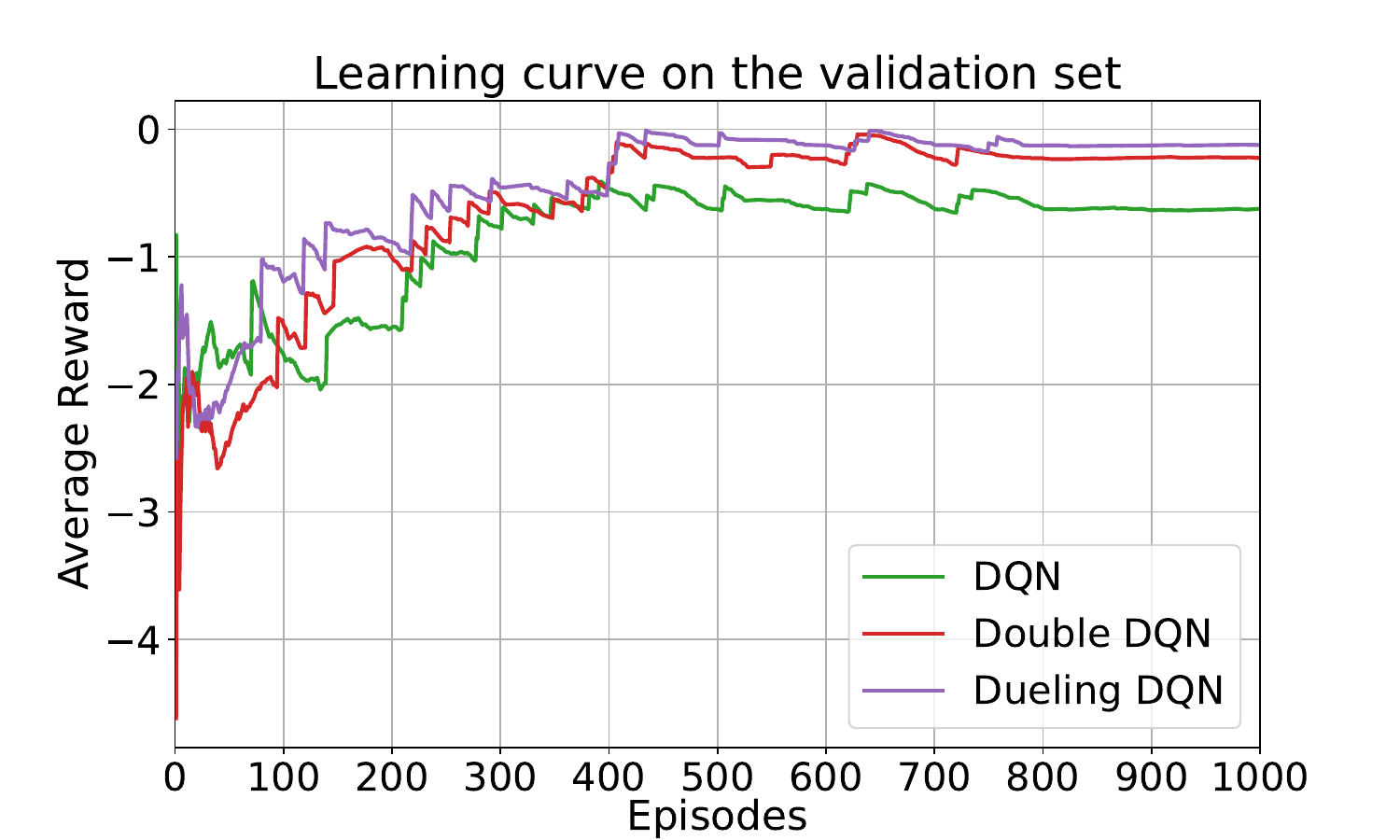}}
    \hspace{-50pt}
    \caption{Learning curves showing the convergence of the different DQN methods, demonstrating that the suggested Dueling DQN algorithm yields the highest average episode rewards.}
    \label{fig:episode_results}
\end{figure*}

\begin{figure*}[!t]
\vspace{-15pt}
    \centering
    \hspace{-50pt}
    \subfloat[LSTM]{\includegraphics[width=0.38\textwidth]{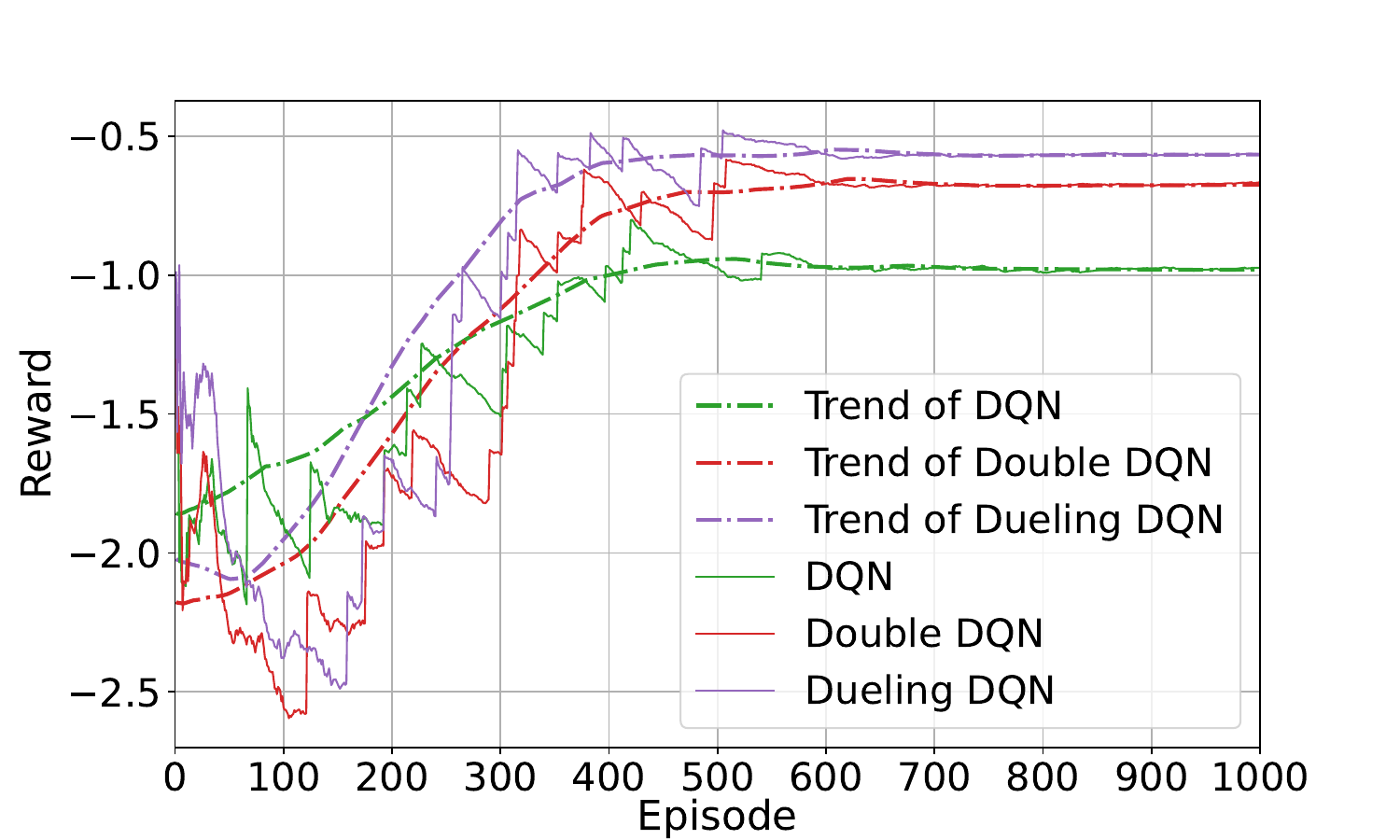}}
    \hspace{-16pt}
    \subfloat[Transformers]{\includegraphics[width=0.38\textwidth]{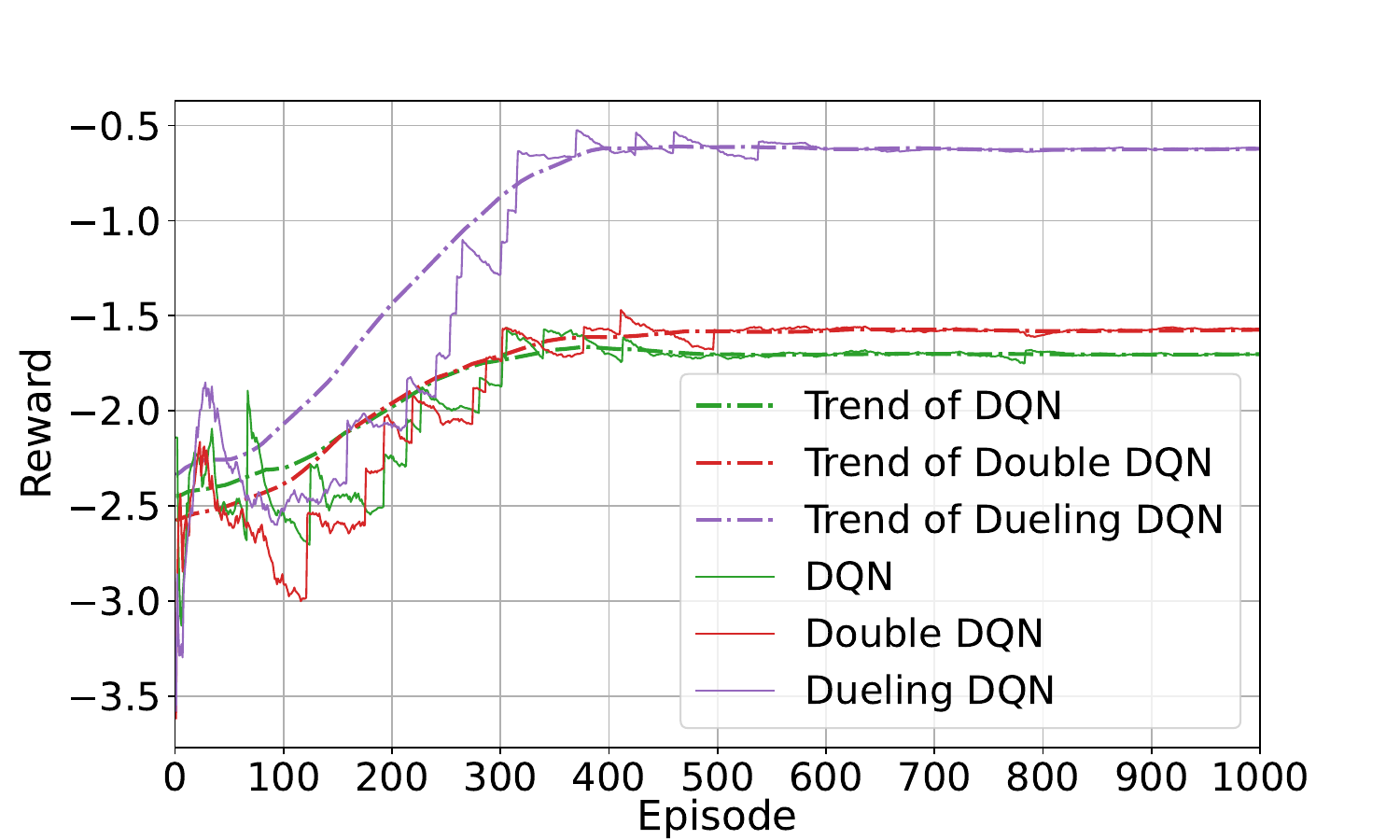}}
    \hspace{-16pt}
    \subfloat[TFT]{\includegraphics[width=0.38\textwidth]{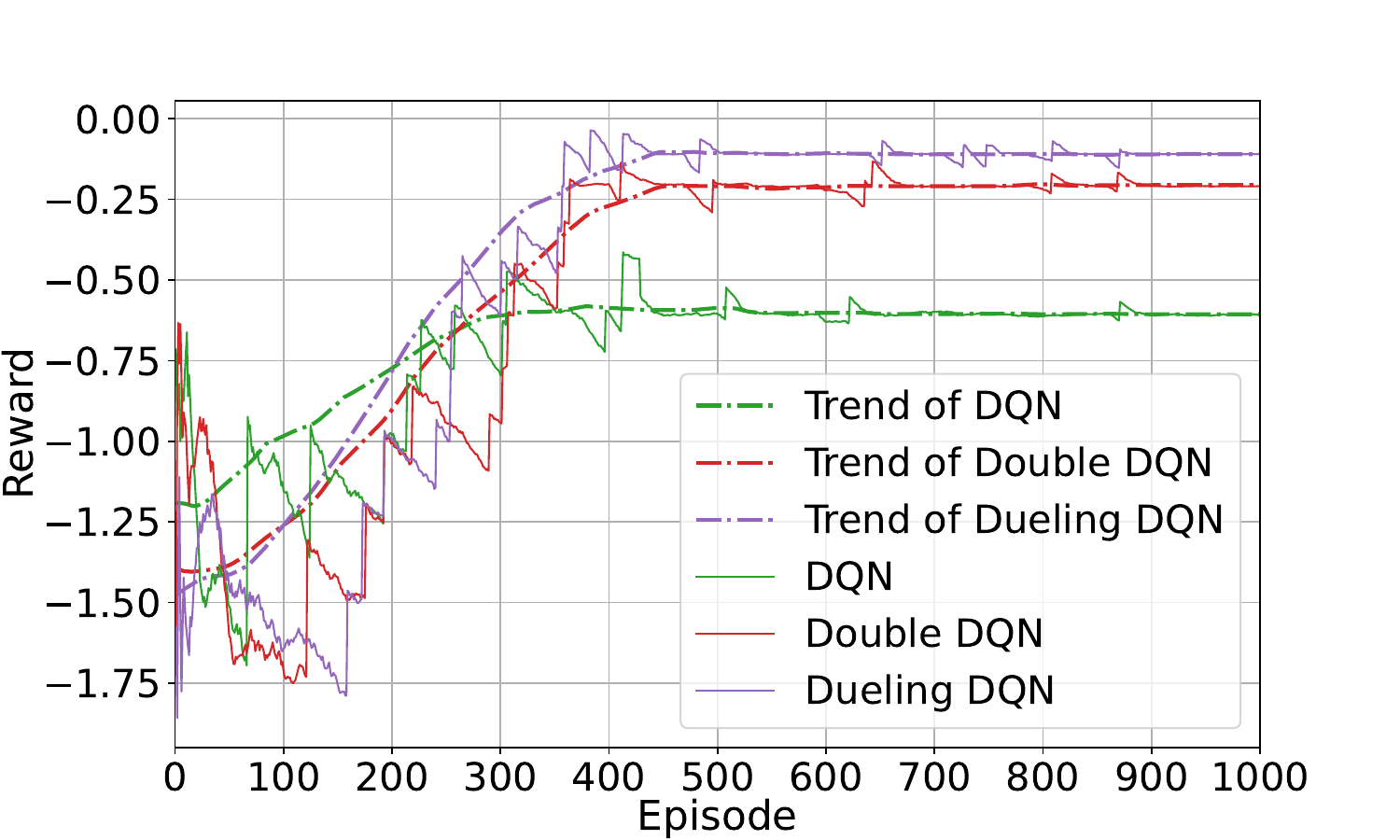}}
    \hspace{-50pt}
    \caption{Agent rewards reveal convergence trends, with dotted lines indicating the reward trend flattening and the solid line depicting actual reward values.}
    \label{fig:episode_results_trend}
\end{figure*}

Benchmark Approaches. The following benchmark approaches are selected for comparison with our approach. The baseline is the "No Policy" (Random Policy) scenario, where actions are chosen randomly without any guidance or learning. In this scenario, the vehicle $v_h$ requests $S_0$ and selects the minimum price in $S_0$, a methodology observed in similar studies such as \cite{zang2019filling,chen2020edge, zang2021soar}. The time interval [$T_0$, $T_f$], in our proposed approach is random and can take different values. In contrast, in most studies, such as the one in \cite{al2022optimal}, where this interval is fixed. The complexity of our problem led us to adopt a multi-phase approach (training and testing). This allows the model to adapt and refine its performance over successive training phases. In contrast, studies such as those in \cite{chen2020edge, al2022optimal} have examined single phase approaches, providing a basis for comparison with the more nuanced multi-phasing method.

\subsection{Experimental Results}

In reinforcement learning, the focus is on the learning curve of the agent performance over time in the environment, which is usually measured through various metrics such as average reward or episode return.

As depicted in Fig. \ref{fig:episode_results}, the learning curve shows the convergence of the average cumulative reward during training episodes for an agent trained to handle various prices, BS numbers, and reservation times. The agent, which uses LSTM, Transformers, and TFT estimation models, converges to the optimal policy after approximately 800 episodes for LSTM and Transformers, and about 600 episodes for TFT. This demonstrates the efficiency of TFT as a learning approach for solving the bandwidth reservation problem.

In particular, Dueling DQN outperforms both DQN and Double DQN in terms of average reward convergence across all three estimation models. This superior performance of Dueling DQN can be attributed to its unique architecture that separates the value and advantage streams, leading to improved generalization, especially in complex and high-dimensional state spaces.

Among the three models (LSTM, Transformers, and TFT), TFT achieves the highest average reward. This can be attributed to TFT’s design for tasks with sequential data and its strong ability to model temporal dependencies. The hybrid architecture of TFT combines LSTM-like recurrent components with the self-attention mechanism found in Transformers, enabling it to retain essential information over time and efficiently model sequential data.

\begin{figure*}[t]
    \centering
    \hspace{-50pt}
    \subfloat[LSTM]{\includegraphics[width=0.38\textwidth]{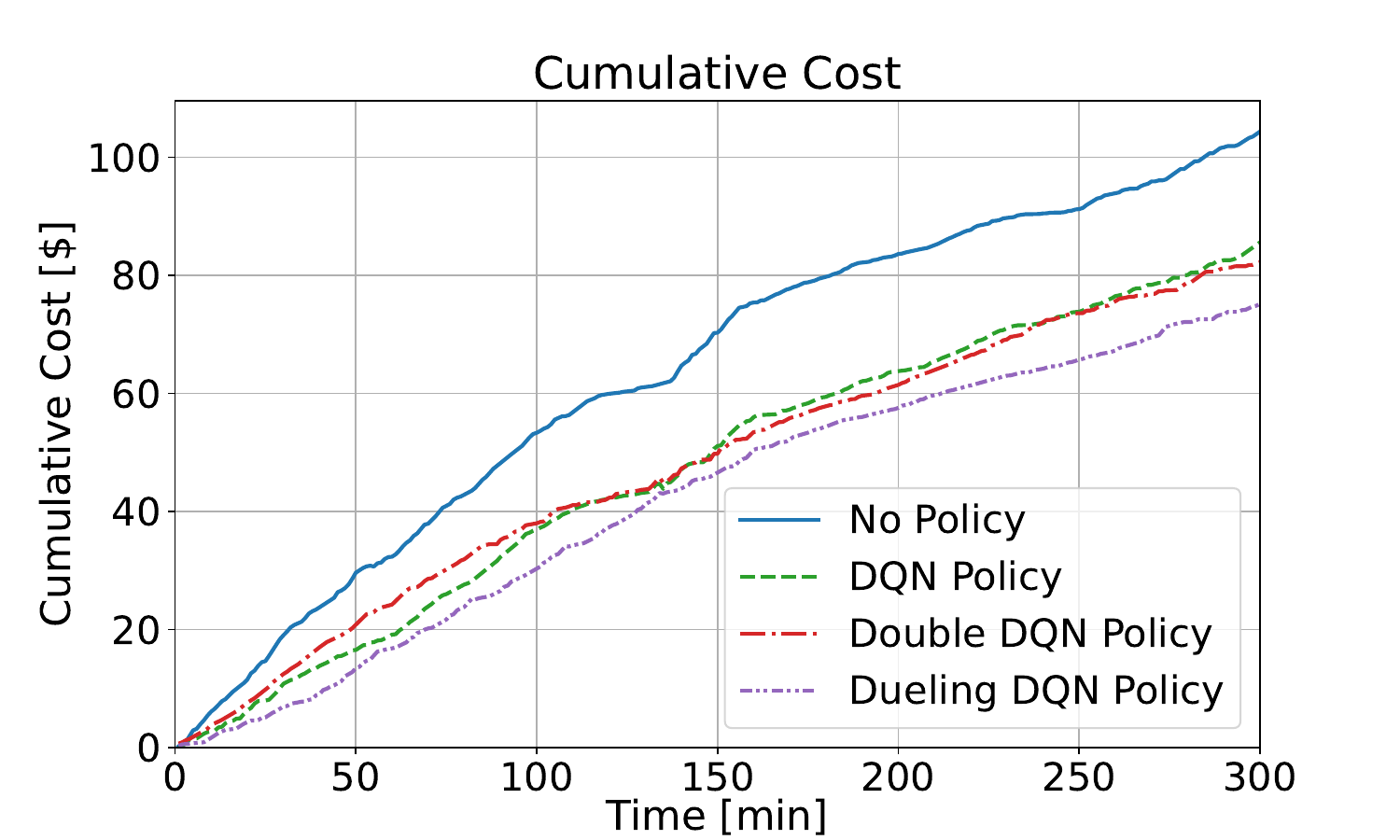}}
    \hspace{-16pt}
    \subfloat[Transformers]{\includegraphics[width=0.38\textwidth]{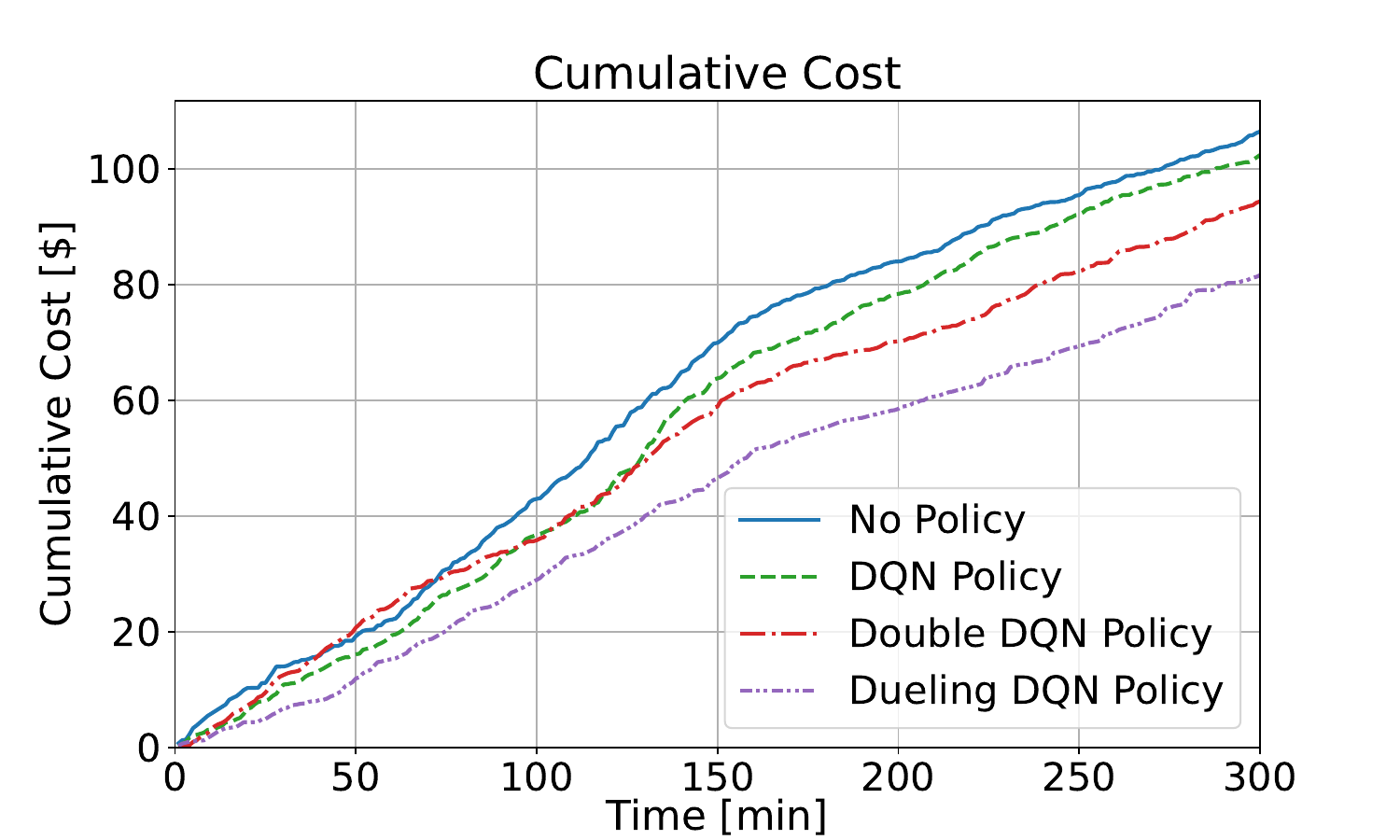}}
    \hspace{-16pt}
    \subfloat[TFT]{\includegraphics[width=0.38\textwidth]{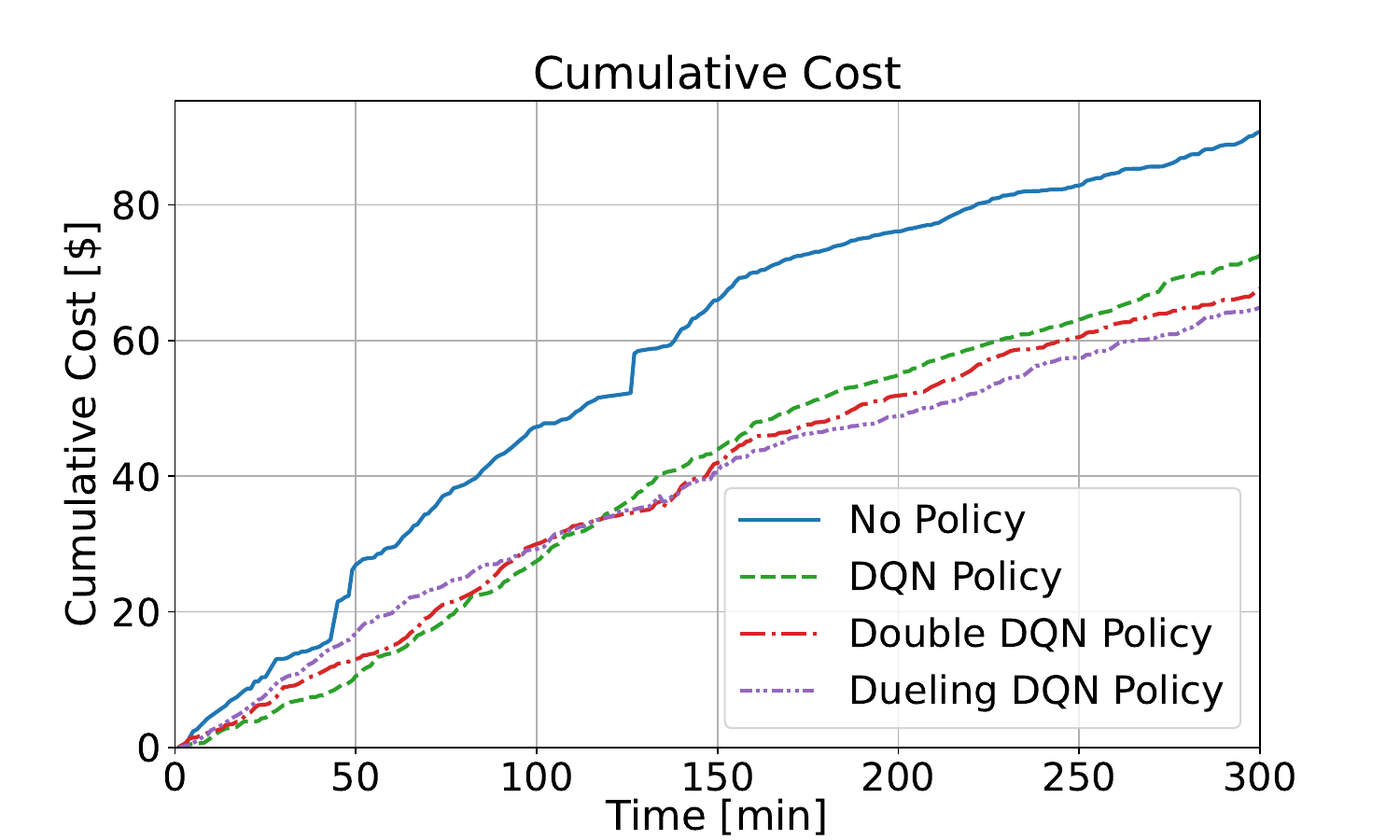}}
    \hspace{-50pt}
    \caption{Cumulative episode rewards are significantly lower when using the Dueling Deep Q-Network (DQN) compared to other methods.}
    \label{fig:accumulate_cost}
\end{figure*}
\begin{figure*}[t]
\vspace{-10pt}
    \centering
    \hspace{-50pt}
    \subfloat[LSTM]{\includegraphics[width=0.38\textwidth]{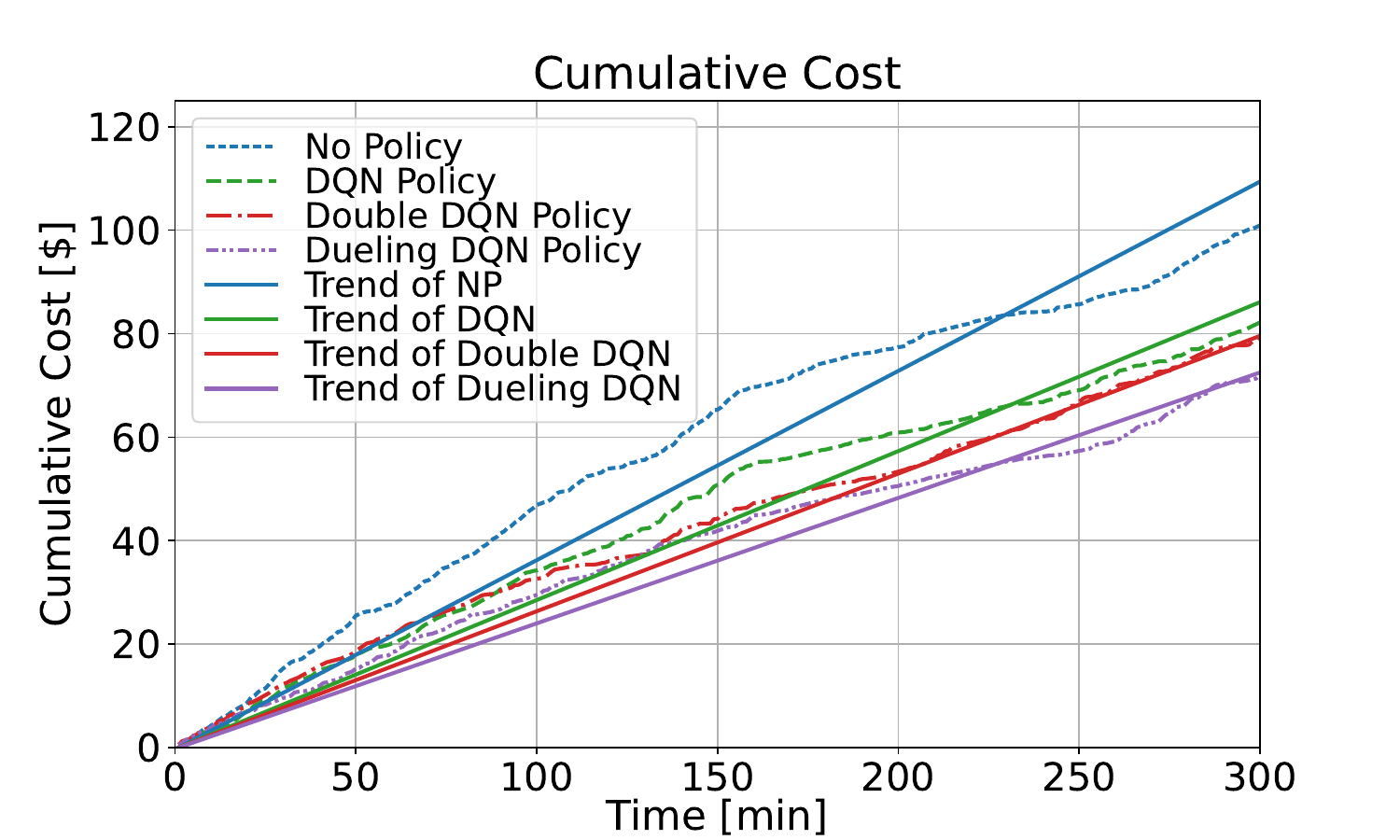}}
    \hspace{-16pt}
    \subfloat[Transformers]{\includegraphics[width=0.38\textwidth]{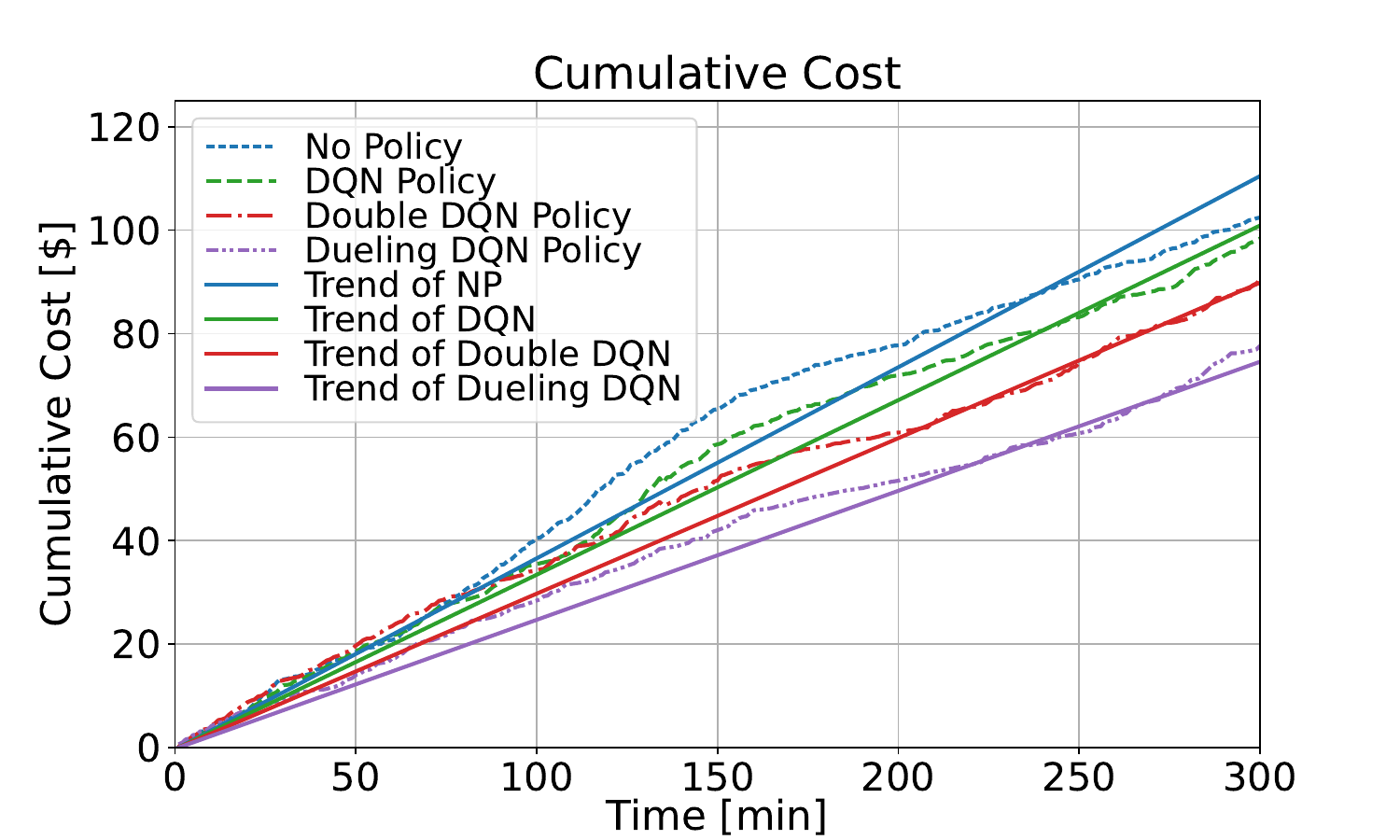}}
    \hspace{-16pt}
    \subfloat[TFT]{\includegraphics[width=0.38\textwidth]{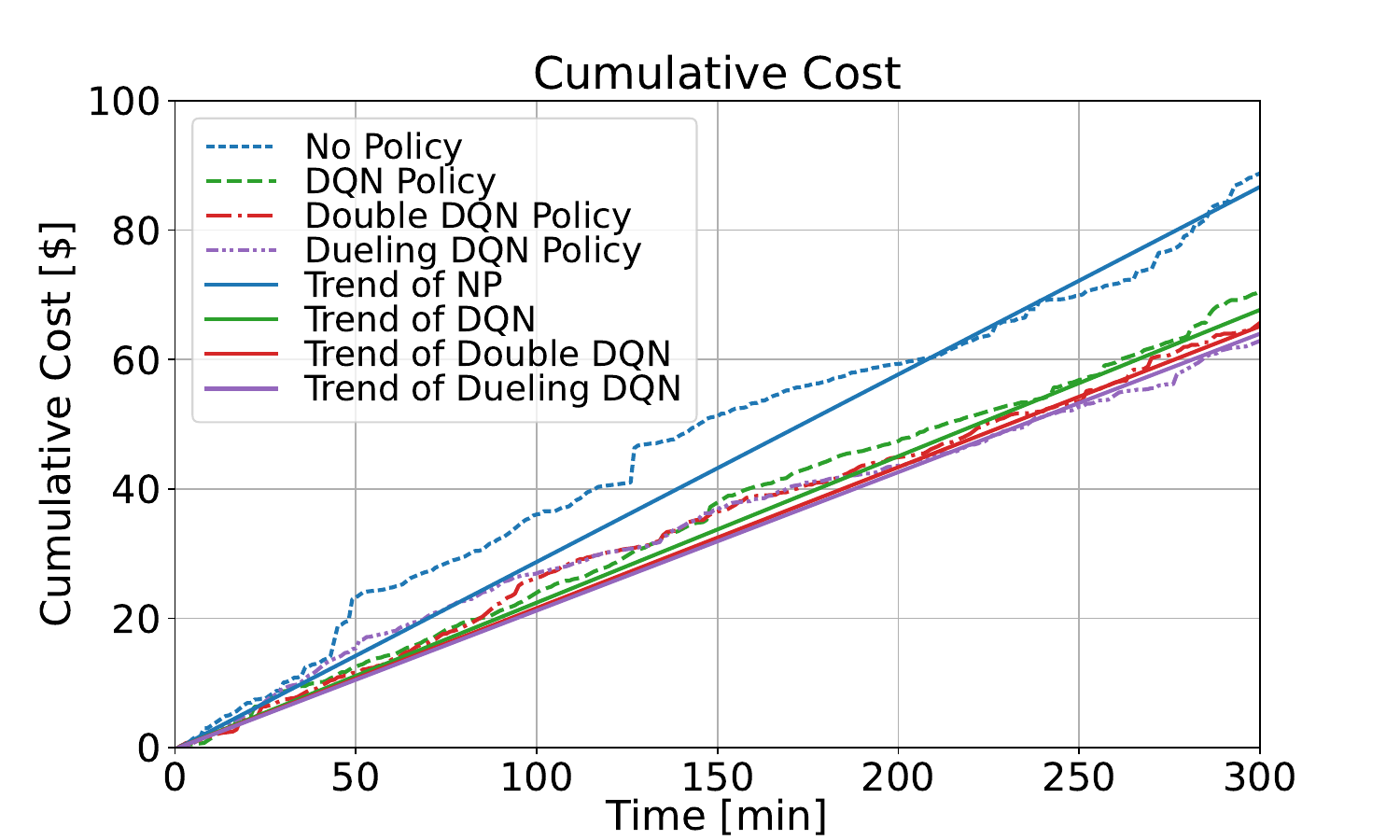}}
    \hspace{-50pt}
    \caption{Cumulative reservation costs show that the Dueling DQN policy outperforms alternative approaches. The solid lines show the trend in cumulative reservation costs, while the dashed line shows the actual cumulative costs.}
    \label{fig:accumulate_cost_trend}
\end{figure*}

To evaluate the model performance, the dataset is divided into training and test datasets. As shown in Fig. \ref{fig:episode_results_trend}, we analyze the convergence by examining the reward-episode graph. In this graph, the dotted lines represent the trend of the reward, which should ideally flatten out, while the solid line shows the actual reward values. The key observation here is to ensure that the reward converges to a stable value. Non-convergence indicates that the agent failed to achieve optimal control, and deploying such an agent can lead to suboptimal actions. In this case, all three models converge after approximately 500 episodes. The absolute value of the reward lacks practical units. It serves as a numerical representation of waiting time and reservation cost, as determined by the reward design outlined in Formula 14. As seen in the learning curve graph, Dueling DQN consistently outperforms DQN and Double DQN because it is associated with the highest rewards. For example, if the LSTM is considered, the convergence values are about -0.97 for DQN, approximately -0.66 for Double DQN, and -0.56 for Dueling DQN. The convergence values for Transformers are, instead, equal to around -1.70 for DQN, approximately -1.57 for Double DQN, and roughly -0.62 for Dueling DQN. For TFT, instead, they are around -0.60 in DQN, approximately -0.20 in Double DQN, and roughly -0.10 in Dueling DQN. By these values it's worth noting that the TFT is associated with the highest reward for all three methods: DQN, Double DQN, and Dueling DQN, hence, it is associated with the best combination of waiting time and reservation cost. These results can be explained similarly to the results observed in the learning curve on the validation set.

Fig. \ref {fig:accumulate_cost} illustrates that the Dueling DQN policy outperforms other approaches (No policy, DQN policy, and Double policy) when considering cumulative reservation costs across all estimation methods (LSTM, Transformers, TFT). The cumulative costs of Dueling DQN policy at time 300 are minimized when compared to the other three methods (No policy, DQN policy, and Double policy). As a result, using this method allows the agent to achieve cost savings. These policies represent different strategies for reinforcement learning agents. "No policy" implies random actions with no learning, while "DQN," "Double DQN," and "Dueling DQN" policies involve deep reinforcement learning techniques to make more informed action choices. Each has its own advantages and is suited for different problem domains and challenges. In no policy (Random policy), actions are chosen randomly without any guidance or learning, and it's often used as a baseline or for comparison purposes to evaluate the performance of other policies such as \cite{zang2019filling,chen2020edge, zang2021soar}. In traditional DQN the policy is based on Q-values, where the agent learns to estimate the expected cumulative future rewards for different actions and chooses the action with the highest estimated Q-value. Double DQN involves using two separate networks to estimate the Q-values, which can lead to more stable and accurate action selection. Instead, Dueling DQN allows the agent to evaluate the state and actions independently, potentially improving learning efficiency and convergence. When Dueling DQN is evaluated across the different estimation methods (LSTM, TFT, and Transformers), it becomes clear that at time 300, TFT stands out as the cost-saving option, linked to the lowest cumulative cost.

\begin{figure*}[t]
    \centering
    \hspace{-50pt}
    \subfloat[LSTM]{\includegraphics[width=0.38\textwidth]{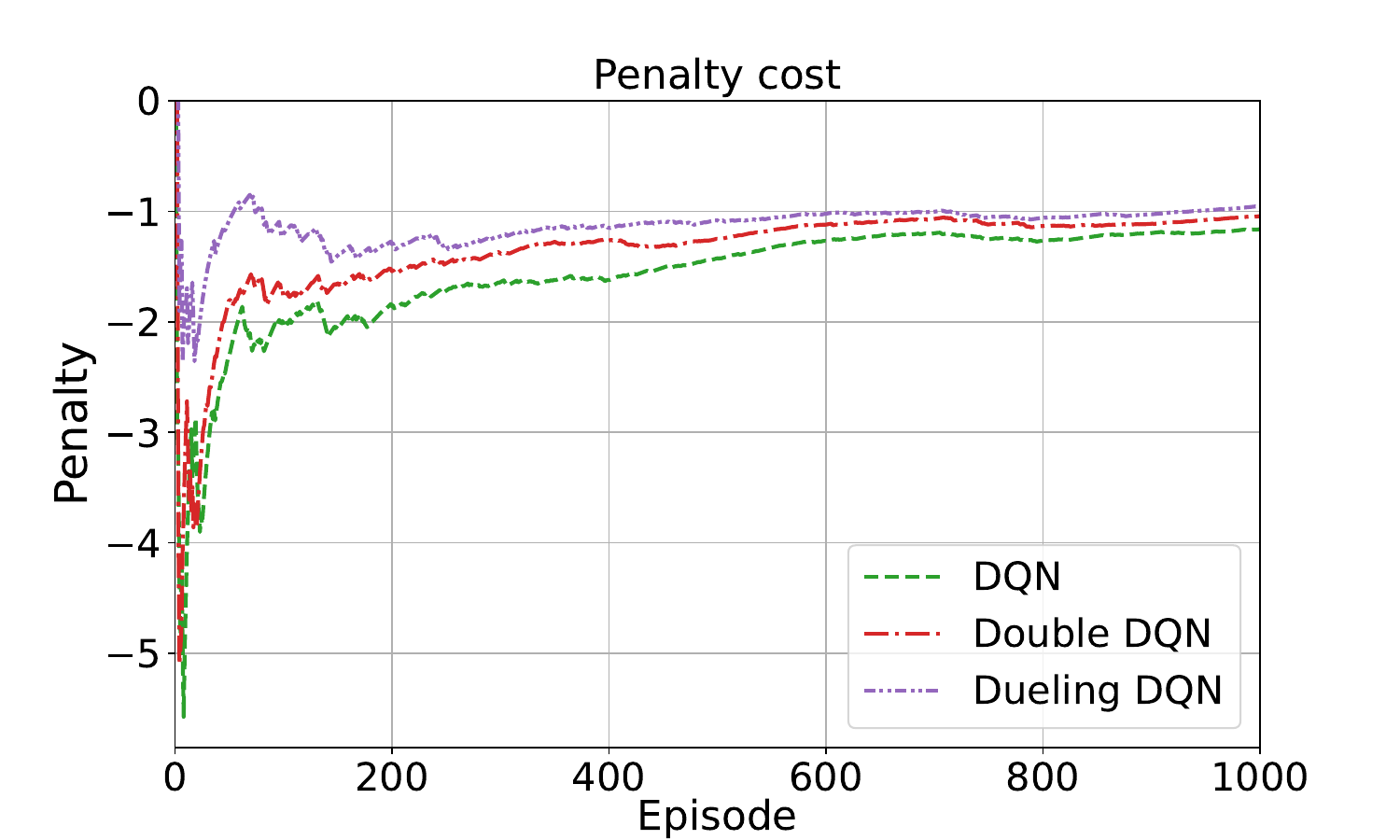}}
    \hspace{-16pt}
    \subfloat[Transformers]{\includegraphics[width=0.38\textwidth]{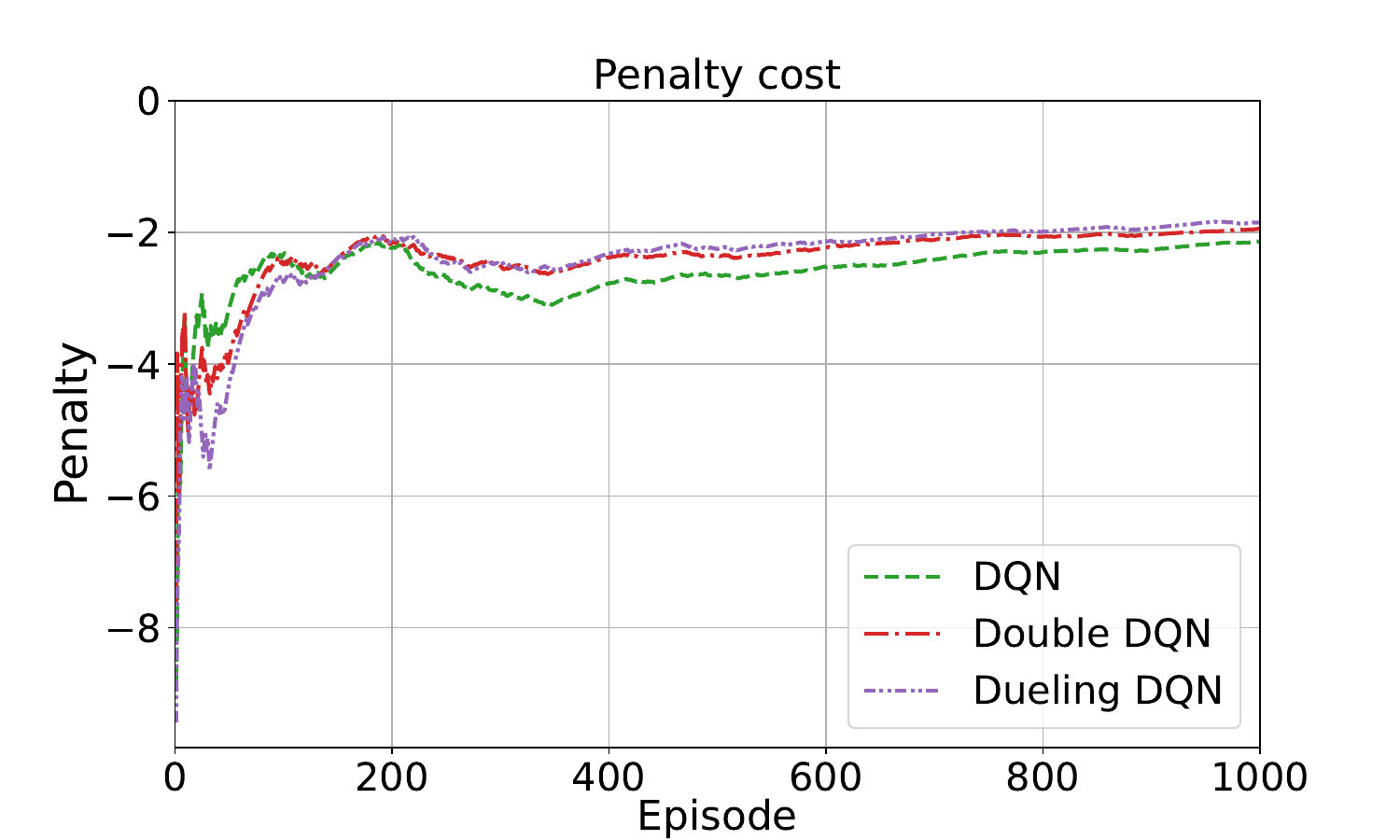}}
    \hspace{-16pt}
    \subfloat[TFT]{\includegraphics[width=0.38\textwidth]{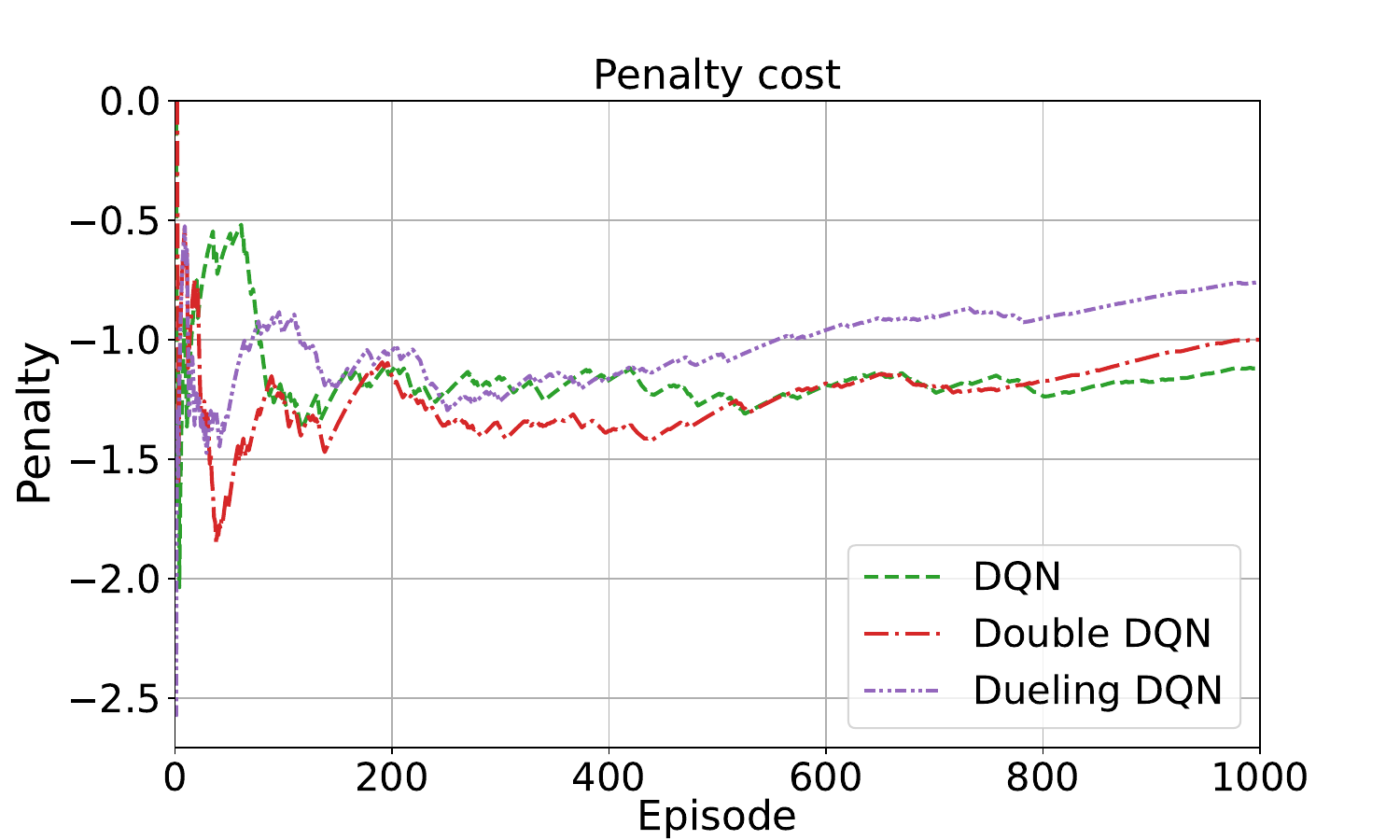}}
    \hspace{-50pt}
    \caption{The penalty cost shows that the most effective penalty value is associated with Dueling DQN, especially in the context of TFT with Dueling DQN.}
    \label{fig:penalty_cost}
\end{figure*}

\begin{figure}[t]
 \centering
\begin{minipage}[t]{0.75\linewidth}
    \includegraphics[width=6.1cm, height=3.8cm]{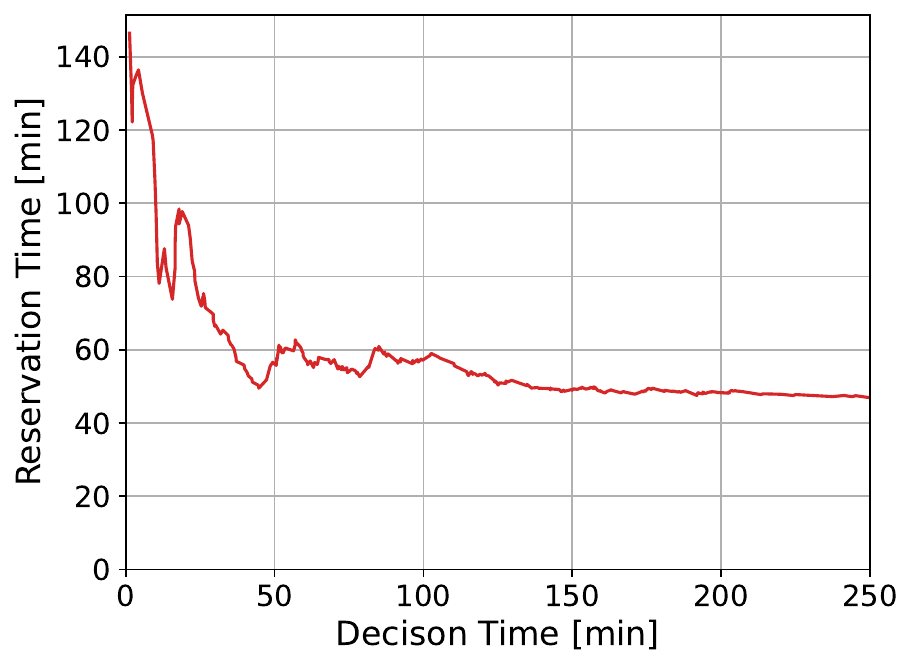}
    
\end{minipage}%
\vspace{-4pt}
\caption{The relation between decision time and reservation time in Dueling DQN with TFT indicates that as decision time increases, reservation time tends to decrease.}
\label{fig:decision_times}
\end{figure}

\begin{figure}[t]
\centering
\vspace{-5pt}
{\includegraphics[width=0.38\textwidth]{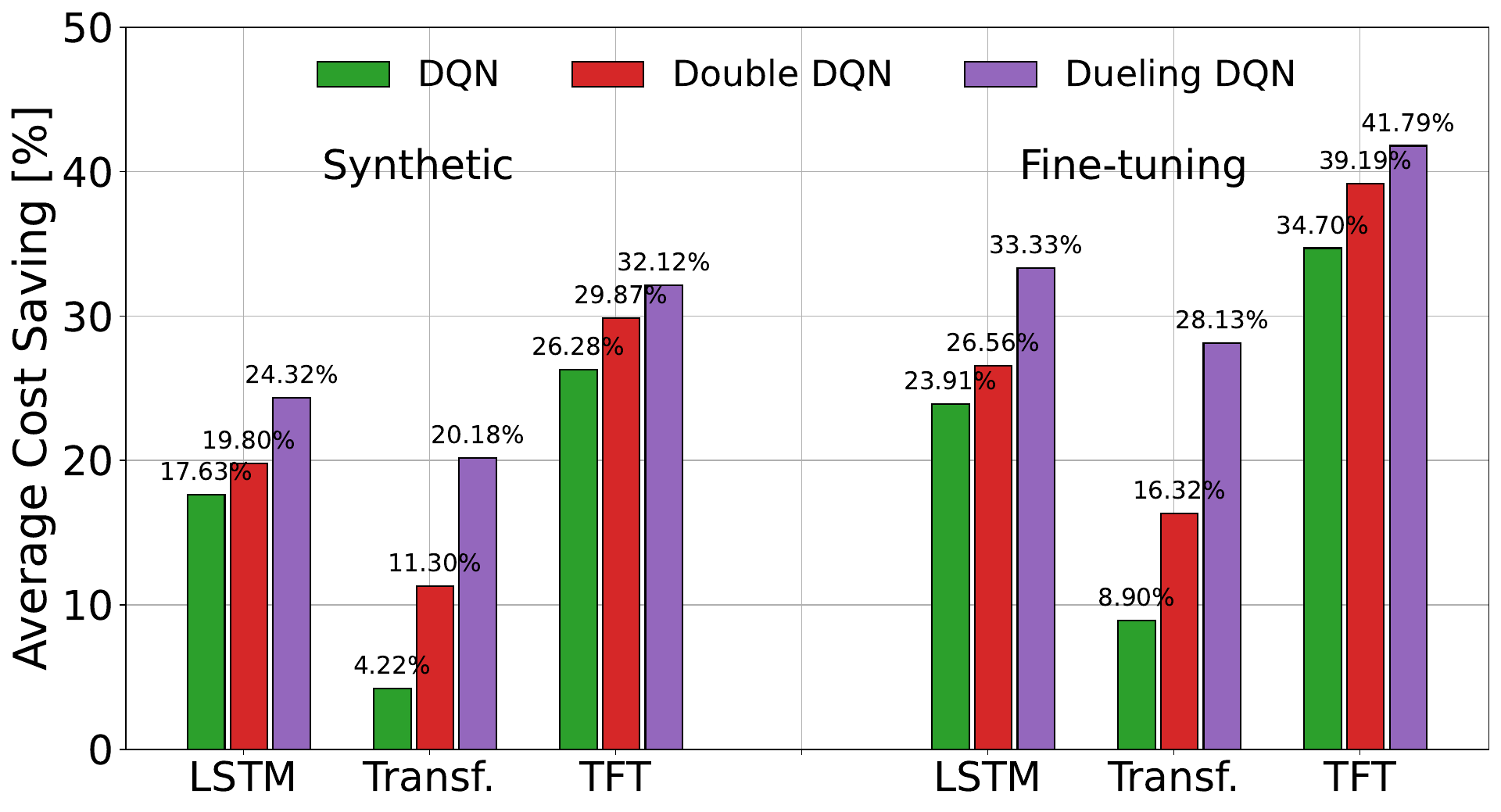}}
\caption{Average cost savings of synthetic and fine-tuned models using various estimation models and DRL methods on test data.}
\label{fig:cost_save1}
\end{figure}

\begin{figure}[t]
\centering
\vspace{-2pt}
{\includegraphics[width=0.38\textwidth]{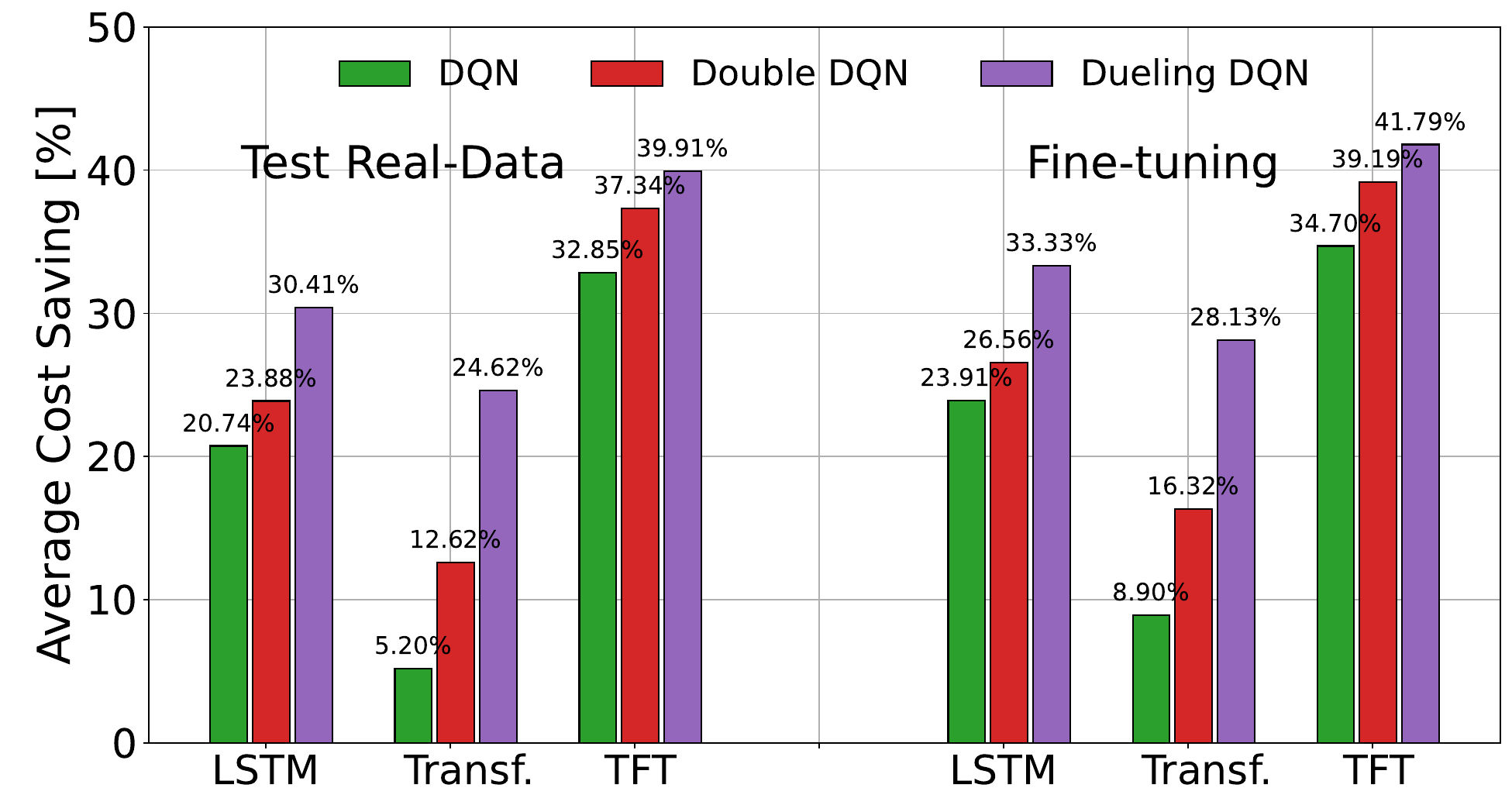}}
\caption{Average cost savings of the fine-tuned model tested on recent data and the training test-data for estimation models using DRL methods.}
\label{fig:cost_save2}
\end{figure}

\begin{figure}[t]
\centering

{\includegraphics[width=0.38\textwidth]{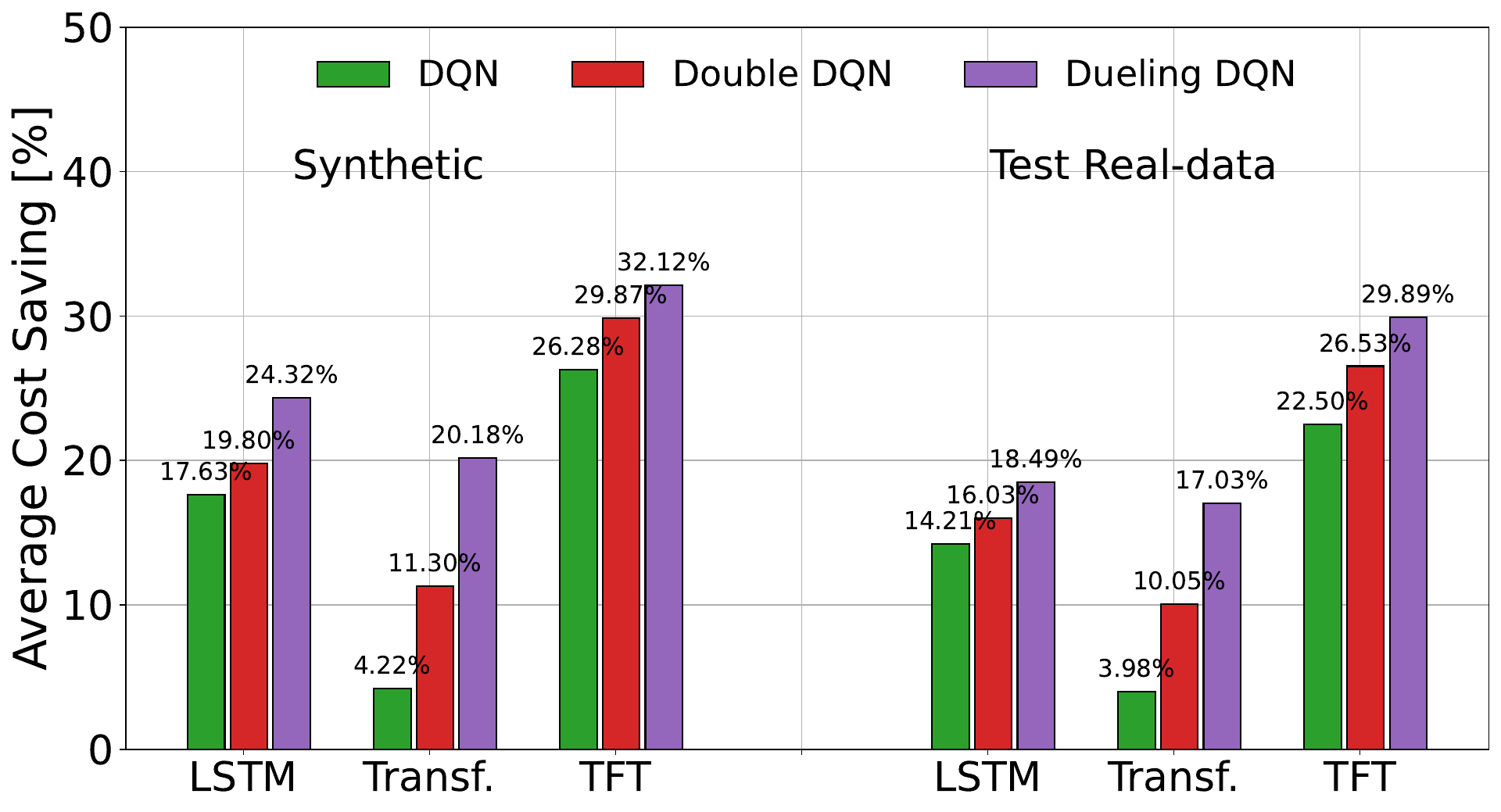}}
 
\caption{Average cost savings of the synthetic model tested on recent data and the training test-data for estimation models using DRL methods.}
\label{fig:cost_save3}
\end{figure}

In Fig. \ref {fig:accumulate_cost_trend}, the trend of cumulative costs is displayed, facilitating a more precise evaluation of the pattern over time. The dotted line corresponds to cumulative costs, while the solid line represents the linear trend. It is important to note that the linear trend for Dueling DQN consistently remains below the trend lines of the other three models (No policy, DQN policy, Double DQN policy). This implies that the Dueling DQN policy provides cost savings by consistently matching the lowest available prices. To compare the different models (LSTM, Transformers, TFT), the analysis focuses on the angle coefficient of the Dueling DQN policy trend relative to each. The angle coefficient for LSTM is approximately 0.24, Transformers approximately 0.25, and for TFT approximately 0.21. As a result, TFT emerges as the most favorable model, consistently positioning it below the other linear trends over time given its smallest angle coefficient. This implies that TFT with the Dueling DQN policy allows for cost savings by consistently targeting the lowest available prices.

Penalty costs represent the impact on the agent of waiting time, as shown in Formula 14. When evaluating different methods (DQN, Double DQN, and Dueling DQN) and different estimation models (LSTM, Transformers, and TFT), it is important to consider that the superior method or model is associated with the lowest absolute value of the penalty. Since penalties are consistently negative, the smallest absolute value corresponds to the highest value. Examining Fig. \ref {fig:penalty_cost}, a comparison of DQN, Double DQN, and Dueling DQN in episode 600 shows that the optimal penalty value is attributed to Dueling DQN, specifically to TFT with Dueling DQN.

The Fig. \ref{fig:decision_times} illustrates that an increase in decision time ($\tau_D$) results in a decrease in reservation request time. The vehicle determines the $\tau_D$ as the point at which it makes a reservation request to the MNOs. The model assists the vehicle in selecting the reservation time based on $\tau_D$. If the vehicle has sufficient $\tau_D$, it can minimize the reservation time. Furthermore, the $\tau_D$ precedes the actual reservation time. For example, if the $\tau_D$ is 10 minutes and the model selects a reservation time after 100 minutes, it means that the departure time starts after 100 minutes.

Fig. \ref{fig:cost_save1} presents the average cost savings achieved by various estimation models (LSTM, Transformers, TFT) and different DRL methods (DQN, Double DQN, Dueling DQN). The calculation involves a comparison between the average cost incurred using the model from phase 1 of the algorithm (synthetic model) and the model from phase 2 (fine-tuned model), tested on the test dataset of the fine-tuned model.

We observe that the highest average cost savings are achieved by the TFT Dueling DQN architecture in both models. The fine-tuned model achieves higher cost savings due to the model's ability to adapt its policy based on the fine-tuning process.  This process refines the model parameters to better match the nuances of the specific test dataset. The observed increase in cost savings highlights the effectiveness of employing transfer learning techniques, specifically fine-tuning.
This observation is further supported by comparing Fig.\ref{fig:cost_save2} and Fig.\ref{fig:cost_save3}. When the models are tested with the most recent data from this year, we observe a significant improvement in the fine-tuned model with TFT and Dueling DQN. Specifically, there is about a 10\% increase in cost savings, calculated as (39.91-29.89)\%, compared to the synthetic model. This underscores the critical importance of the fine-tuning phase in the performance of the model.

\section{CONCLUSION}
In conclusion, this research effectively addresses resource provisioning challenges, especially in the context of safety-critical vehicular applications and bandwidth reservation requests. The approach to determine the optimal prices in the context of a multi-MNO scenario, has been observed as an instance of the secretary problem, which has proven highly effective in addressing these challenges using the Dueling Deep Q-Learning algorithm and utilizing the TFT for training the DRL agent before letting it take the decision of making reservations through the utilization of real price data and multi-phase training, resulting in significant cost reductions by 40 \% in edge environments, as we have provided in the experimental results. In future work, we plan to explore methods that require less hyperparameter tuning, a common requirement for conventional DQN models. Additionally, we intend to enhance the pre-trained transformers, which show promise in scenarios with limited data. In addition, we aim to extend our framework to incorporate multi-agent scenarios where collaboration among vehicles or between vehicles and infrastructure could lead to further optimization opportunities. This involve adapting MARL algorithms to function effectively under the computational constraints of in-vehicle hardware and ensuring efficient communication between agents. Finally, we will enhance the reward function by incorporating network latency and reliability metrics for a balanced optimization of cost, latency, and reliability. We also plan to investigate QoS-aware strategies, multi-objective optimization techniques, and real-world testing to validate performance in dynamic vehicular network scenarios.


\end{document}